\documentclass[sigconf]{aamas} 

\usepackage{algorithm}
\usepackage{algorithmic}
\usepackage{xcolor}
\usepackage{thmtools}
\usepackage{thm-restate}
\newtheorem{theorem}{Theorem}

\newtheorem{lemma}[theorem]{Lemma}

\newtheorem{definition}[theorem]{Definition}

\usepackage{balance} 

\usepackage{newfloat}
\usepackage{listings}
\DeclareCaptionStyle{ruled}{labelfont=normalfont,labelsep=colon,strut=off} 
\DeclareCaptionStyle{ruled}{labelfont=normalfont,labelsep=colon,strut=off} 
\floatstyle{ruled}
\newfloat{listing}{tb}{lst}{}
\floatname{listing}{Listing}

\setcopyright{ifaamas}
\acmConference[AAMAS '24]{Proc.\@ of the 23rd International Conference
on Autonomous Agents and Multiagent Systems (AAMAS 2024)}{May 6 -- 10, 2024}
{Auckland, New Zealand}{N.~Alechina, V.~Dignum, M.~Dastani, J.S.~Sichman (eds.)}
\copyrightyear{2024}
\acmYear{2024}
\acmDOI{}
\acmPrice{}
\acmISBN{}

\acmSubmissionID{1021}

\title[AAMAS-2024 Formatting Instructions]{Policy Learning for Off-Dynamics RL with Deficient Support}

\author{Linh Le Pham Van}
\affiliation{
  \institution{Applied Artificial Intelligence Institute, Deakin University}
  \city{Geelong}
  \country{Australia}}
\email{l.le@deakin.edu.au}

\author{Hung The Tran}
\affiliation{
  \institution{Applied Artificial Intelligence Institute, Deakin University}
  \city{Geelong}
  \country{Australia}}
\email{hung.tranthe@deakin.edu.au}

\author{Sunil Gupta}
\affiliation{
  \institution{Applied Artificial Intelligence Institute, Deakin University}
  \city{Geelong}
  \country{Australia}}
\email{sunil.gupta@deakin.edu.au}

\begin{abstract}
Reinforcement Learning (RL) can effectively learn complex policies. However, learning these policies often demands extensive trial-and-error interactions with the environment. In many real-world scenarios, this approach is not practical due to the high costs of data collection and safety concerns. As a result, a common strategy is to transfer a policy trained in a low-cost, rapid source simulator to a real-world target environment. However, this process poses challenges. Simulators, no matter how advanced, cannot perfectly replicate the intricacies of the real world, leading to dynamics discrepancies between the source and target environments. Past research posited that the source domain must encompass all possible target transitions, a condition we term full support. However, expecting full support is often unrealistic, especially in scenarios where significant dynamics discrepancies arise. In this paper, our emphasis shifts to addressing large dynamics mismatch adaptation. We move away from the stringent full support condition of earlier research, focusing instead on crafting an effective policy for the target domain. Our proposed approach is simple but effective. It is anchored in the central concepts of the skewing and extension of source support towards target support to mitigate support deficiencies. Through comprehensive testing on a varied set of benchmarks, our method's efficacy stands out, showcasing notable improvements over previous techniques.
\end{abstract}


\newcommand{\BibTeX}{\rm B\kern-.05em{\sc i\kern-.025em b}\kern-.08em\TeX}

\makeatletter
\gdef\@copyrightpermission{
	\begin{minipage}{0.3\columnwidth}
		\href{https://creativecommons.org/licenses/by/4.0/}{\includegraphics[width=0.90\textwidth]{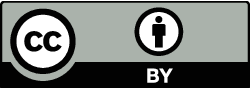}}
	\end{minipage}\hfill
	\begin{minipage}{0.7\columnwidth}
		\href{https://creativecommons.org/licenses/by/4.0/}{This work is licensed under a Creative Commons Attribution International 4.0 License.}
	\end{minipage}
	\vspace{5pt}
}
\makeatother

\begin{document}


\pagestyle{fancy}
\fancyhead{}


\maketitle 


\section{Introduction}

Reinforcement Learning (RL) has shown its capacity to acquire intricate behaviors in numerous real-world challenges \cite{mnih2015human, levine2016end, schrittwieser2020mastering}. However, it requires numerous trial-and-error interactions with the environment, which may not be feasible due to the high costs of data collection or safety concerns in many real-world scenarios (e.g. robotics, autonomous driving, medical treatment, etc). 
Training the policy in an alternative source environment, e.g. a simulator, which is both safer and faster, and using a limited set of data from the real-world target environment has, therefore, become a common approach. This approach is usually known as Off-Dynamics Reinforcement Learning \cite{eysenbach2020off}.

Previous work on this problem includes \cite{ljung1998system, chebotar2019closing, peng2018sim, andrychowicz2020learning}. \citet{ljung1998system} and \citet{chebotar2019closing} propose an approach to align the source dynamics to the target dynamics by real-world data using the system identification method. These methods require a detailed understanding of the target domain (e.g. knowing the physics behind the systems). \citet{peng2018sim} train the policy on a set of randomized simulators to yield a robust policy. Again, the set of randomizations is chosen carefully based on the detailed understanding of the target domain. The requirement of having a detailed understanding of target and source domains limits the applicability of such methods. \emph{Thus new methods that do not rely on such detailed knowledge are required}.

The other recent works such as \cite{hanna2017grounded, desai2020stochastic, karnan2020reinforced, desai2020imitation} learn the target policy via learning an action transformation function which maps the actions suggested by the source policy to make them suitable for the target domain. In a related approach, \citet{eysenbach2020off, liu2022dara} use the dynamics discrepancy term as an additional reward to prevent the policy from exploiting the dynamics mismatch area. However, all these works make a strong assumption that the source domain (e.g. a simulator) encompasses all possible target (real-world)  transitions, a condition which we call \emph{Full support}. However, full support condition rarely holds in practice as a simulator no matter how advanced cannot perfectly replicate the intricacies of the real world, and thus can not cover all the transitions in the target domain. For example, an autonomous driving vehicle may face changed conditions such as new kind of places (i.e. highway, city, countryside), weather (i.e. sunny, rainy, hazy), or time (i.e. day, night), resulting in only a fraction of target transitions in the support of the source domain. In Section \ref{sec:experiment_section}, we show that the existing methods fail drastically when a source domain does not fully support the target domain. \emph{Therefore, the problem of off-dynamics reinforcement learning under deficient support remains an open problem}.

In this paper, we address the aforementioned challenges relaxing both the detailed domain understanding and full support requirements in Off-dynamics RL to deal with source support deficiency. Under this setting, we propose an effective method to reduce source deficiency by creating a \emph{modified source} domain using two operations: (1) by \emph{skewing} the source transitions to support the target domain with higher probability, (2) and \emph{extending} the source transitions towards the target transitions to improve the source support for the target. The skewing is guided by an importance weighting which is learned by solving an optimization problem and the source support extension is done by following the MixUp scheme \citep{zhang2017mixup}. Finally, we utilize both the skewed and mixup transitions and adjust the rewards to compensate for the dynamics discrepancy between the modified source domain and the target domain, and use this modified data for target policy learning.  

Our main contributions are:
\begin{itemize}
    \item We are the first to address the policy learning for off-dynamics RL with deficient support, which is a novel problem and is encountered in numerous real-world scenarios.
    \item We conduct a theoretical analysis of off-policy RL under the setting of deficent support, offering valuable practical insights for the development of effective policy learning algorithm.
    \item We propose DADS, a practical and effective algorithm for off-dynamics RL with deficient support via source skewing and extension operations.
    \item Finally, we demonstrate the superior performance of our proposed method over the existing methods through a diverse set of experiments.
\end{itemize}
\section{Related Work}
\textbf{Domain adaptation in RL:}
Domain adaptation in RL is needed when there is a difference in the observation space, transition dynamics, or reward function. In this paper, we study domain adaptation with dynamics mismatch. System identification method \cite{ljung1998system, kolev2015physically, yu2017preparing,  chebotar2019closing} is a direct approach to align the source dynamics with the observed target data. However, these methods typically require a model of the source environment and a large set of target data to adjust the parameters of the source environment to align it with the target domain. Another approach, domain randomization \cite{peng2018sim, andrychowicz2020learning, sadeghi2016cad2rl, tobin2017domain}, involves training RL policies over a collection of randomized simulated source domains. However, this approach often exhibits sensitivity to the selection of randomized parameters or parameter distributions \cite{eysenbach2020off}.

In contrast, ground action transformation techniques \cite{hanna2017grounded, karnan2020reinforced, desai2020stochastic, desai2020imitation, zhang2021policy} eliminate the need for a parameterized simulator or manually selected randomized dynamics parameters. These techniques aim to rectify dynamics mismatch by learning action transformations of the source policy using the target data. Such action transformation techniques require the existence of an accurate action transformation policy, which may be infeasible when the dynamics mismatch between the source and target is large e.g. when the source domain lacks transitions seen in the target domain. In the absence of an accurate action transformation policy, such methods exhibit poor performance. 

\citet{xu2023cross} introduce value-guided data filtering that removes the transitions that have high value discrepancies during policy training. Recently, \citet{eysenbach2020off, liu2022dara} proposed using dynamics discrepancy to correct the reward during training the policy. However, these works rely on a full support condition, that the source domain must contain all possible transitions in the target domain, which rarely holds in real-world scenarios, thus preventing handling the large dynamics gap problems. Our study takes advantage of the reward correction but relaxes the full support condition. 

\textbf{Mixup in RL:}
MixUp was first introduced in \cite{zhang2017mixup} in the supervised learning setting as a novel data augmentation method that improves the generalizability of the deep learning models by training them on convex linear combinations of dataset samples. In the RL problems, \cite{wang2020improving, sander2022neighborhood, lin2021continuous, zhang2022generalization} have demonstrated that MixUp helps to improve the generalizability and sample efficiency of the learned policy. From a different perspective, we employ MixUp to expand the support of the source domain to cover the support of the target domain, reducing the support deficiency problem.

\textbf{Deficient support in Off-policy Bandits:}
The deficient support problem has been explored in off-policy bandit settings, where it refers to the lack of data for certain actions under the logging policy compared to the target policy \cite{sachdeva2020off, tran2021combining, felicioni2022off, saito2023off}. However, this existing work focuses primarily on the policy space. In contrast, our work tackles the novel challenge of deficient support in the context of off-dynamics policy learning for reinforcement learning. We address the gap between the source and target transition dynamics distributions, which is a critical issue for effective policy adaptation in RL.

\section{Problem Setting and Preliminaries}
\begin{figure*}
    \centering
    \includegraphics[width=2.1\columnwidth]{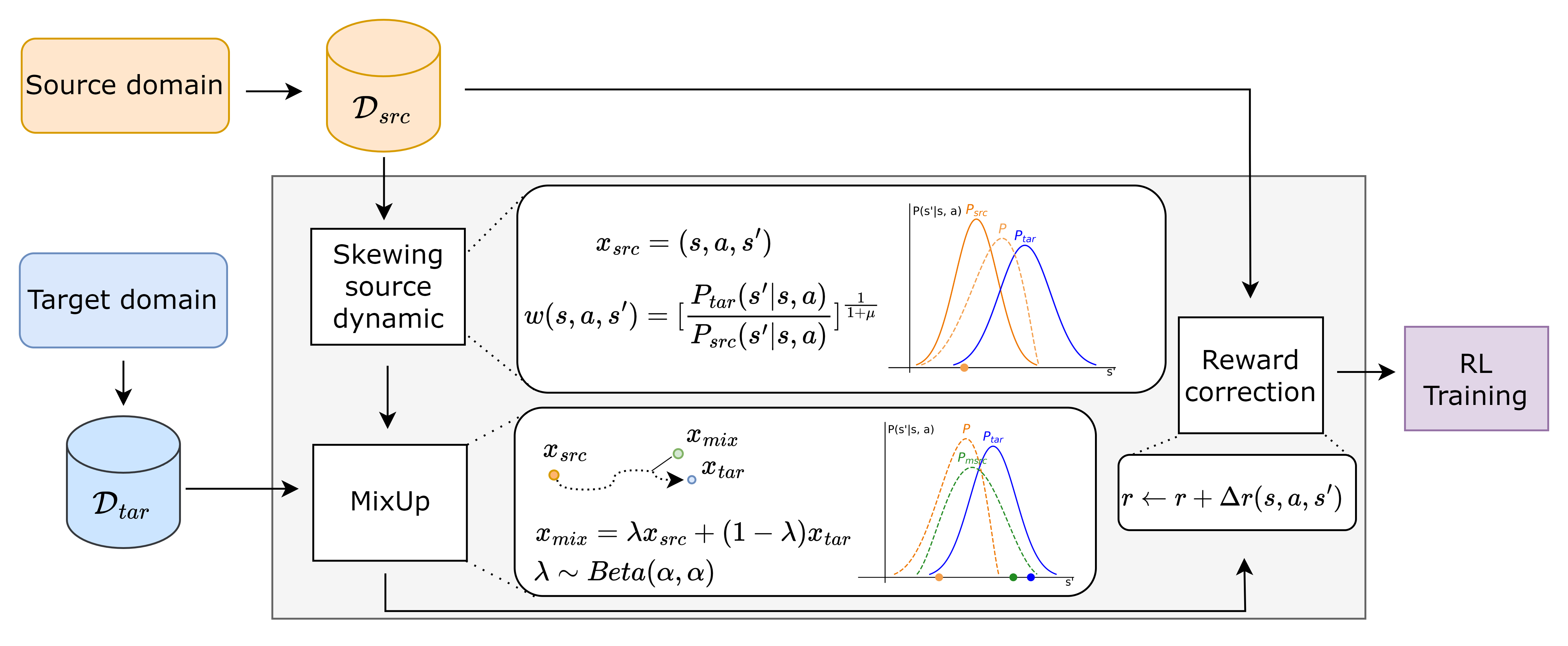}
    \caption{Off-dynamics online policy learning with deficient support. Given the source domain and target domain with limited online interaction, we propose skewing the source dynamics, which enables us to sample the source transitions that are closely aligned with the target dynamics and employ MixUp procedure to expand the source support set towards the target support set. Then we adopt the Reward correction to compensate the policy with an additional reward for encouraging dynamic-consistent behaviors.}
    \label{fig:method}
\end{figure*}
\textbf{Background:}
In this section, we introduce our notation and a formal definition of off-dynamics online policy learning. We consider two infinite-horizon Markov Decision Processes (MDPs) $\mathcal{M}_{src} := (\mathcal{S}, \mathcal{A}, P_{src}, r, \gamma, \rho_0)$ and $\mathcal{M}_{tar} := (\mathcal{S}, \mathcal{A}, P_{tar}, r, \gamma, \rho_0)$ representing source domain and target domain, respectively. In our setting, we assume that the two domains share the same state space $\mathcal{S}$, action space $\mathcal{A}$, reward function $r: \mathcal{S} \times \mathcal{A} \rightarrow \mathbb{R}$, discount factor $\gamma \in [0, 1)$, and the initial state distribution $\rho_0 : \mathcal{S} \rightarrow [0, 1]$; the only difference between two domains is in their transition dynamics, $P_{src}(s'|s, a)$ and $P_{tar}(s'|s, a)$.

A policy $\pi : \mathcal{S} \rightarrow \mathcal{P}(\mathcal{A})$ is defined as a map from states to a probability distribution over actions. Then, we denote the probability that the policy $\pi$ encounters state $s$ at the time step $t$ in an MDP $\mathcal{M}$ as $\mathrm{P}^{\pi}_{\mathcal{M}, t}(s)$, 
and the normalized state-action occupancy of state-action pair $(s, a)$ in $\mathcal{M}$ is $\rho^{\pi}_{\mathcal{M}}(s, a) := (1 - \gamma)\sum^{\infty}_{t=0}\mathrm{P}^{\pi}_{\mathcal{M}, t}(s)\pi(a|s)$. The performance of a policy $\pi$ in the MDP $\mathcal{M}$ is defined as $\eta_{\mathcal{M}}(\pi) = \mathbb{E}_{s, a \sim \rho_{\mathcal{M}}^\pi}[r(s, a)]$. The value function on the MDP $\mathcal{M}$ and policy $\pi$ is defined as $V_{\mathcal{M}}^\pi(s):=\underset{\pi, P}{\mathbb{E}}\left[\sum_{t=0}^{\infty} \gamma^t r\left(s_t, a_t\right) \mid s_0=s\right]$.

We focus on the Off-Dynamics Online (ODO) policy learning, which is formally defined as follows:
\begin{definition}[Off-Dynamics Online Policy Learning]
Given a source domain represented by $\mathcal{M}_{src}$ and a target domain represented by $\mathcal{M}_{tar}$ with distinct dynamics functions, our goal is to leverage source interactions and a small amount of target interactions to derive a good policy that achieves high reward in $\mathcal{M}_{tar}$.
\end{definition}

We highlight that the previous methods often requires the assumption of a \textit{full support condition}, that implies every possible transition in the target domain $\mathcal{M}_{tar}$ is covered by the source domain $\mathcal{M}_{src}$. We formally define the \textit{full support condition} as follows:
\begin{definition}[Full support]
We say that a source domain $\mathcal{M}_{src}$ has full support for a target domain $\mathcal{M}_{tar}$ if every transition with non-zero probability in the target domain $\mathcal{M}_{tar}$ also has a non-zero probability in the source domain $\mathcal{M}_{src}$:
$P_{tar}(s'|s, a) > 0 \Rightarrow P_{src}(s'|s, a) > 0, \forall s, s' \in \mathcal{S}, a \in \mathcal{A}.$
\end{definition}

In the ODO policy learning  with the \textit{full support} condition holds, the target performance can be guaranteed as follows:
\begin{restatable}[Performance bound]{proposition}{progapfull}
\label{pro_gap_1}
Let $\mathcal{M}_{src}$ and $\mathcal{M}_{tar}$ are the source domain and target domain with different dynamics $P_{src}$ and $P_{tar}$ respectively. The performance difference of any policy $\pi$ in $\mathcal{M}_{src}$ and $\mathcal{M}_{tar}$ can be bounded as follows:
    \begin{equation}
    \begin{aligned}
        &\left|\eta_{M_{tar}}(\pi) -\eta_{M_{src}}(\pi)\right| \\
        &\leq\frac{\gamma r_{max}}{(1-\gamma)^2}\cdot\underbrace{\sqrt{2\mathbb{E}_{\rho_{tar}^\pi}\left[D_{KL}(P_{tar}(.|s, a), P_{src}(.|s, a))\right]}}_{(a)}.
    \end{aligned}
    \label{full_support_gap_eq}
\end{equation}
\end{restatable}

The performance bound, as outlined in Proposition \ref{pro_gap_1}, depends on the dynamics discrepancy term $(a)$. A recent approach called DARC \citep{eysenbach2020off} uses the dynamics discrepancy between source and target domains as an incremental reward, to prevent the policy from exploiting areas in the source that have a high dynamics mismatch with the target domain.

However, the \emph{full support condition} is stringent and might not hold in many real-world scenarios.
When this condition is not met, it results in the challenge of \textit{deficient support}, which we formally define as:
\begin{definition}[Deficient support]
    We say source MDP $\mathcal{M}_{src}$ has support deficiency for target MDP $\mathcal{M}_{tar}$ if there exists a set $\{(s', s, a)\} \neq \emptyset$ such that for each transition $(s', s, a)$ belongs to it, we have $P_{tar}(s'|s, a) > 0$ but $P_{src}(s'|s, a) = 0$. 
\end{definition}

The deficient support assumption does not require the source domain to encompass all potential target transitions.
Thus, when deficient support happens, it poses a challenge due to the uncovered target areas. In this paper, we relax the \textit{full support} assumption and propose a method for the off-dynamics online (ODO) policy learning with \emph{deficient support}. 

Under the deficient support problem, we derive the performance bound as follows:
\begin{restatable}[Performance bound under Defficient Support]{proposition}{progapdefficient}
\label{pro_gap_2}
Let $\mathcal{M}_{src}$ and $\mathcal{M}_{tar}$ are source domain and target domain with different dynamics $P_{src}$ and $P_{tar}$ respectively. For each state-action pair $s, a$, denote $S_{s, a}^0 = \{[s'_{0_i}, s'_{0_j}]\}$ contains intervals where $P_{src}(.|s, a) = 0$, and $S_{s, a}^1 = \{[s'_{1_i}, s'_{1_j}]\}$ includes intervals where $P_{src}(.|s, a) > 0$, and $S_{s, a}^0 \cup S_{s, a}^1 = supp(P_{tar}(.|s, a))$.
The performance difference of any policy $\pi$ in $\mathcal{M}_{src}$ and $\mathcal{M}_{tar}$ can be bounded as follows:
   
    \begin{equation}
    \begin{aligned}
        &\left|\eta_{\mathcal{M}_{tar}}(\pi) - \eta_{\mathcal{M}_{src}}(\pi)\right| \\
        &\leq \frac{\gamma r_{max}}{(1-\gamma)^2} \cdot\mathbb{E}_{\rho_{tar}^\pi(s, a)}\left[\sum_{S_{s, a}^1}\left|\int_{s'_{1_i}}^{s'_{1_j}} P_{tar}(s'|s, a) - P_{src}(s'|s,a)ds'\right|\right] \\
        &+\frac{\gamma}{1 - \gamma}\cdot \underbrace{\mathbb{E}_{\rho_{tar}^\pi(s, a)}\left[\sum_{S_{s, a}^0}\int_{s'_{0_i}}^{s'_{0_j}} P_{tar}(s'|s, a)\cdot \left|V_{src}^\pi(s')\right|ds'\right]}_{\text{support deficiency}}.
    \end{aligned}
    \label{deficient_support_gap_eq}
\end{equation}
\end{restatable}
Proposition \ref{pro_gap_2} highlights the gap between $\eta_{\mathcal{M}_{tar}}(\pi)$ and $\eta_{\mathcal{M}_{src}}(\pi)$ due to support deficiency.
Notably, the In Eq (\ref{full_support_gap_eq}) emerges as a special instance of In Eq (\ref{deficient_support_gap_eq}) when full support is assumed. In this special case, the support deficiency term in (\ref{deficient_support_gap_eq}), which quantifies the target value $V_{tar}$ on the unsupported set, vanishes. Based on Proposition \ref{pro_gap_2}, guaranteeing performance on the target domain hinges on minimizing both the dynamics discrepancy within the supported region and the support deficiency. With this dual objective in mind, the next section introduces our method aimed at simultaneously reducing both terms to ensure robust performance guarantees.

\section{Proposed Method}
In this section, we introduce a novel approach to address the challenge of off-dynamics policy learning in the presence of support deficiency. Our method aims to create a modified source domain that has minimum source deficiency w.r.t to the target. Our method has three primary steps: (1) skewing the source transitions to maximize its support overlap with the target domain; (2) extrapolating the source transitions to extend the source support all the way up to the target domain. This is done using the MixUp procedure by creating new synthetic transitions between the source transitions and the target transitions via their convex combinations; and (3) combining the source and MixUp transitions to form a modified source transition set, adjusting their rewards similarly to \cite{eysenbach2020off}, and use these modified transitions to train the target policy. Our method is depicted in Figure \ref{fig:method}. 
Our approach aims to minimize the second and third terms in the performance bound in In Eq (\ref{deficient_support_gap_eq}). The second term  is minimised by iteratively \emph{skewing} the source support towards the target. The third term in the performance bound is minimized by \emph{extending} the source support toward the target via mix up as it can generate the samples in the unsupported region.

\subsection{Skewing Source Dynamics} \label{ssec:skew_source}
We present the skewing source dynamics strategy, which enables us to sample the source transitions that are closely aligned with the target dynamics. 
Specifically, we learn a dynamics distribution $P(s'|s, a)$ that is close to the target dynamics distribution $P_{tar}(s'|s, a)$ but not significantly far from the source dynamics $P_{src}(s'|s, a)$, measured in terms of the KL divergence. This is formulated as the following constrained function optimization problem:
\begin{equation}
\begin{aligned} \label{eq:dynamic_optimization_first}
    &\min_{P \in \mathcal{P}} D_{\mathrm{KL}}(P(.|s,a)||P_{tar}(.|s,a)) \text{ } \forall s \in \mathcal{S}, a \in \mathcal{A}. \\
    & \text{s.t } D_{\mathrm{KL}}(P(.|s,a)||P_{src}(.|s,a)) \leq \epsilon \text{ } \forall s \in \mathcal{S}, a \in \mathcal{A}. \\
    & \int_{s'} P(s'|s, a)ds' = 1 \text{ } \forall s \in \mathcal{S}, a \in \mathcal{A}.
\end{aligned}    
\end{equation}
where $\mathcal{P}$ is a family of all transition dynamics distributions and we have $P_{tar}(.|s,a)\in \mathcal{P}$ and $P_{src}(.|s,a)\in \mathcal{P}$. The first constraint with KL divergence and the parameter $\epsilon$ regularizes the dynamics function $P(s'|s,a)$ to stay close to the source dynamics $P_{src}(s'|s, a)$, while the second constraint is to ensure $P(s'|s, a)$ is a valid probability density function. 

To solve the constrained optimization problem, we create a Lagrangian-based objective function and then solve for the optimal transition dynamics $P^*(s'|s, a)$ by taking a derivative and equating it to zero. With a few steps of analysis, we obtain the optimal skewed dynamics function $P^*(s'|s, a)$ as follows:
\begin{equation}\label{eq:p_optimal_2}
\begin{aligned}
    P^*(s'|s,a) &\propto P_{src}(s'|s, a)\exp[\frac{1}{1 + \mu}\log \frac{P_{tar}(s'|s,a)}
    {P_{src}(s'|s,a)}].
\end{aligned}
\end{equation}
The parameter $\mu$ serves as the Lagrange multiplier linked to the KL constraint in (\ref{eq:dynamic_optimization_first}), and it essentially dictates the degree of constraint strength or how far the optimal solution can deviate from the source.
The full derivation is provided in the \emph{Appendix}.

To effectively sample the skewed distribution $P^*$, we leverage the existing source domain data $P_{src}$. We achieve this by re-weighting samples from $P_{src}$ based on the density ratio between the skewed transition dynamics and the original (source) transition dynamics. In particular, we propose a sampling approach where source transitions $(s, a, s')$ are chosen with probabilities proportional to the density ratio $w(s, a, s')$ as follows:
\begin{equation}
\begin{split}
    w(s, a, s') & = \exp\left[\frac{1}{1 + \mu}(\log \frac{P_{tar}(s'|s,a)}{P_{src}(s'|s,a)})\right] \\ &\propto P^*(s'|s, a) / P_{src}(s'|s, a).
\end{split}
\end{equation}
Concretely, we define the sampling probability of the $i$-th source transition $(s, a, s')$ as follows: 
\begin{equation}\label{eq:probability_sample}
    p^i(s, a, s') = \frac{w^i(s, a, s')}{\sum_k w^k(s, a, s')}.
\end{equation}
where $w^i(s, a, s')$ is the priority weight of source transition $i$. 

\textbf{Estimating the ratio of source and target transition dynamics:} \label{par:mix}
Calculating the sampling probability of each source transition requires estimating the density ratio between the target dynamics and the source dynamics for each source transition. Similar to \cite{eysenbach2020off}, we adopt the probabilistic classification technique \cite{sugiyama2012density} to estimate this density ratio. Specifically, we use a pair of binary classifiers, $q_{\theta_{SAS}}(.|s, a, s')$ and $q_{\theta_{SA}}(.|s, a)$, which distinguish whether a transition $(s,a,s')$ (or a state-action pair $(s,a)$) comes from the source or target domain. The density ratio is computed as follows:
\begin{equation}\label{eq:priority_cal}
\begin{aligned}
    \log \frac{P_{tar}(s'|s,a)}{P_{src}(s'|s,a)} = &\log \frac{q_{\theta_{SAS}}(\text{target}|s, a, s')}{ q_{\theta_{SAS}}(\text{source}|s, a, s')} \\ &+ \log \frac{q_{\theta_{SA}}(\text{source}|s, a)}{q_{\theta_{SA}}(\text{target}|s, a)}.  
\end{aligned} 
\end{equation}
The two classifiers $q_{\theta_{SAS}}(.|s, a, s')$ and $q_{\theta_{SA}}(.|s, a)$ are learned with the standard cross-entropy loss using the source and target data. 
\subsection{Extending Source Support}
The deficient support presents the existence of uncovered target areas. Our idea is to extend the source support toward the target support by employing the MixUp procedure, thus filling the uncovered target support. Specifically, we utilize the skewing source transitions from the previous step and mix them up with target transitions to create MixUp transitions. 
While we should ideally be mixing up $s'$ from source and target domains conditioned on the same state-action pair $(s,a)$, since we deal with continuous state and action spaces, it is challenging to find an identical pair. Even a nearest neighbor approach can result in fairly distant $(s,a)$ pairs from the source and target domains. To avoid this problem, we mix up the quadruples $(s,a,r,s')$ of the source with those of the target. 

Given a source transition $x_{src} = (s, a, r, s')_{src}$ (obtained from the skewing step) and a target transition $x_{tar} = (s, a, r, s')_{tar}$, we use MixUp to generate a synthetic transition by taking convex combination between $x_{src}$ and $x_{tar}$ as follows:
\begin{equation}\label{eq:mixup}
    x_{mix} = \lambda x_{src} + (1 - \lambda) x_{tar}
\end{equation}
where $\lambda$ is sampled from a Beta distribution as $\lambda \sim B(\alpha, \alpha), \text{ with } \alpha > 0$. As suggested by \cite{zhang2017mixup}, we set $\alpha = 0.2$. Note that if either source or target transition encounters a terminal state, we do not employ interpolation, and simply use the target transition instead. This is mainly done to avoid non-binary terminal signal \citep{sander2022neighborhood}.

Let $\{x^i_{src}=(s, a, r, s')^i_{src}\}_{i=1}^{N}$ be a batch of $N$ source transitions sampled from the source data $D_{src}$ with probabilities $p^i(s, a, s')$ as in Eq (\ref{eq:probability_sample}). We sample uniformly $N$ target transitions $\{x^i_{tar}=(s, a, r, s')^i_{tar}\}_{i=1}^{N}$ from the target data $D_{tar}$. We then sample a batch $\{\lambda_i\}_{i=1}^{N}$ from a Beta distribution, and perform MixUp using each pair of source and target transitions.

\subsection{Reward modification}
While the modified source now supports the target transitions better, the transition probability densities of the modified source and target domain still may be different. Thus, following the scheme in \cite{eysenbach2020off}, we adjust the reward for each transition by adding an incremental term $\Delta r$, as follows:
\begin{equation}
    \Delta r (s, a, s') = \log P_{tar}(s'|s, a) - \log P_{msrc}(s'|s, a)
\end{equation}
where $P_{msrc}(s'|s, a)$ denotes the transition dynamics of the modified source. In practice, to obtain $\Delta r(s, a, s')$ for each transition $(s, a, s')$, we employ a similar density ratio estimate as in section \ref{ssec:skew_source}, using two binary classifiers $q_{\phi_{SAS}}(.|s, a, s')$ and $q_{\phi_{SA}}(.|s, a)$ as follows:
\begin{equation}\label{eq:delta_r}
\begin{aligned}
    \Delta r(s, a, s') = &\log \frac{q_{\phi_{SAS}}(\text{target}|s, a, s')}{ q_{\phi_{SAS}}(\text{modified source}|s, a, s')} \\ &+ \log \frac{q_{\phi_{SA}}(\text{modified source}|s, a)}{q_{\phi_{SA}}(\text{target}|s, a)}.
\end{aligned}
\end{equation}
Then each transition in the batch of modified source transitions is modified as $(s, a, r + \Delta r, s')$ and used to train the target policy. 
\subsection{Algorithm}
\begin{algorithm}[ht]
\caption{Online Dynamics Adaptation with Deficient Support in RL (DADS)}
\label{alg:mixup}
\begin{algorithmic}[1]\label{alg:proposed}
\STATE\textbf{Input:} Source MDP $\mathcal{M}_{src}$ and target $\mathcal{M}_{tar}$; ratio $r$ of experience from source vs. target; the batch size $N$.
\STATE \textbf{Initialize:} The source data $\mathcal{D}_{src}$ and the target data $\mathcal{D}_{tar}$; policy $\pi$; parameters $\theta$ for classifiers that distinguish source and target domain $q_{\theta_{SAS}}, q_{\theta_{SA}}$; and parameters $\phi$ for classifiers that distinguish modified source and target domain $q_{\phi_{SAS}}, q_{\phi_{SA}}$.
\FOR {$t = 1, \dots, \text{num iterations}$} 
    \STATE $\mathcal{D}_{src} \leftarrow \mathcal{D}_{src} \cup \text{ROLLOUT}(\pi, \mathcal{M}_{src})$. 
    \IF {$t \mod r == 0$} 
        \STATE $\mathcal{D}_{tar} \leftarrow \mathcal{D}_{tar} \cup \text{ROLLOUT}(\pi, \mathcal{M}_{tar})$.
    \ENDIF
    \STATE Sample $N$ source transitions $\{(s, a, r, s')^i_{src}\}^{N}_{i=1}$ with each transition's probability $p^i(s, a, s')$ computed  via Eq (\ref{eq:probability_sample}) from $\mathcal{D}_{src}$ . 
    \STATE Update transition priority in $D_{source}$ via Eq (\ref{eq:priority_cal}).
    \STATE Sample $N$ target transitions $\{(s, a, r, s')^i_{tar}\}^{N}_{i=1}$ uniformly from $\mathcal{D}_{tar}$.
    \STATE Create MixUp transitions $\{(s, a, r, s')^i_{mix}\}^{N}_{i=1} \leftarrow \text{MixUp}(\{(s, a, r, s')^i_{src}\}^{N}_{i=1}, \{(s, a, r, s')^i_{tar}\}^{N}_{i=1})$ via Eq (\ref{eq:mixup}). 
    \STATE Modify the reward for each transition in source and mixup batch with $\Delta r$ via Eq (\ref{eq:delta_r}).
    \STATE Train the source-target classifiers $\theta \leftarrow \theta - \eta\nabla_\theta \text{Cross-entropyLoss}(\mathcal{D}_{src}, \mathcal{D}_{tar}, \theta)$. 
    \STATE Train the modified\_source-target classifiers $\phi \leftarrow \phi - \eta\nabla_\phi \text{Cross-entropyLoss}(\mathcal{D}_{src}, \mathcal{D}_{tar}, \phi)$.
    \STATE Train the policy $\pi$ with modified source and mixup transitions using any standard policy learning algorithm (e.g. SAC).
\ENDFOR
\STATE\textbf{return } $\pi$.
\end{algorithmic}
\end{algorithm}
We summarize the above steps as our proposed method in Algorithm \ref{alg:proposed}. We perform the skew operation in Lines 8 and 9, and the MixUp procedure in  Line 11. We perform reward correction in Line 12 and learn two pair of domain classifiers in Lines 13 and 14. Finally, in Line 15, we use a standard RL algorithm for policy learning. In our experiment, we use Soft Actor-Critic (SAC) \cite{haarnoja2018soft} algorithm. 
To simplify the algorithm, we find that simply using a fixed constant for the Lagrange multiplier $\mu$ is also effective, rather than adaptively updating 
$\mu$. In the implementation, to sample efficiently from Eq (\ref{eq:probability_sample}), we employ the \emph{sum-tree} data structure similar to \cite{schaul2015prioritized}, which allows $O(\log N)$ complexity for updates and sampling.

\textbf{Discussion:} We discuss the key signficance of our two operations: \emph{Skew} and \emph{Extension}.  Skewing source transitions  shifts their probability mass towards the target transitions without changing the source domain's support or generating any new samples. Unlike skewing, extension of source support using MixUp can generate new synthetic samples from outside the source support closer to the target domain. Both schemes have their individual strengths and limitations. In particular, the skewing operation boosts the sampling of source transitions that are close to the target domain. However, these transitions can not fill the uncovered target areas in the source domain.
On the other hand, the MixUp operation can generate novel synthetic samples that can expand the source support to cover unseen target transitions and thus can bring significant improvement in target policy learning.
However, a downside can be that randomly mixing up a target transition with any source transition may lead to synthetic samples that do not lie in the target transition manifold, which could at times degrade the performance. In our case, when we use skewing and MixUp together, skewing helps to improve the performance of MixUp by rejecting the source transitions that are unlikely to occur in the target domain before mixing them. Thus, both skewing and Mixup have independent and effective roles in our algorithm. We empirically demonstrate their effectiveness in Section \ref{sec:experiment_section}.

\section{Experiments}\label{sec:experiment_section}
In this section, we provide an empirical analysis of our proposed approach across different levels of deficient support: small support overlap (small), medium support overlap (medium), and large support overlap (large). Furthermore, through ablation studies, we delve deeper into the significance of each component in our method.
\begin{figure}[H]
    \centering
    \includegraphics[width=1.0\linewidth]{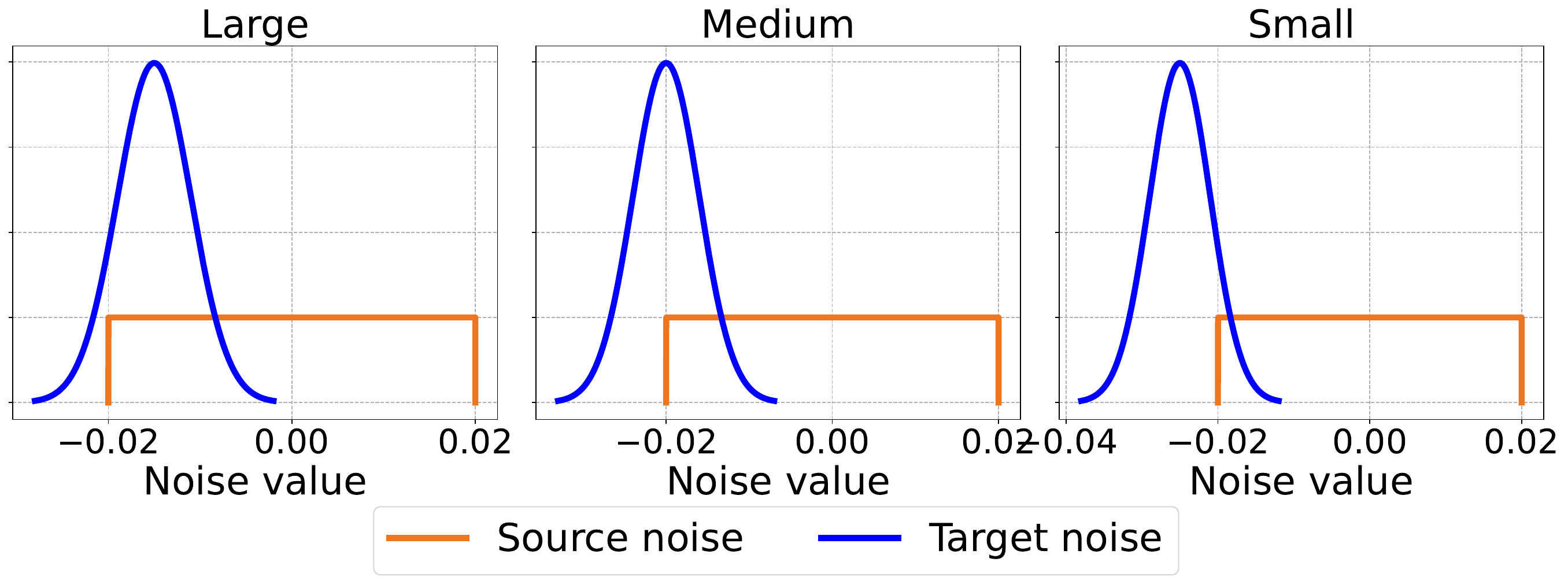}
    \caption{Visualization of the source and target noise distributions in Walker benchmark in three distinct deficient support levels.}
    \label{fig:ds_sample}
\end{figure}
\begin{figure*}
    \centering
     \includegraphics[width=1.0\linewidth]{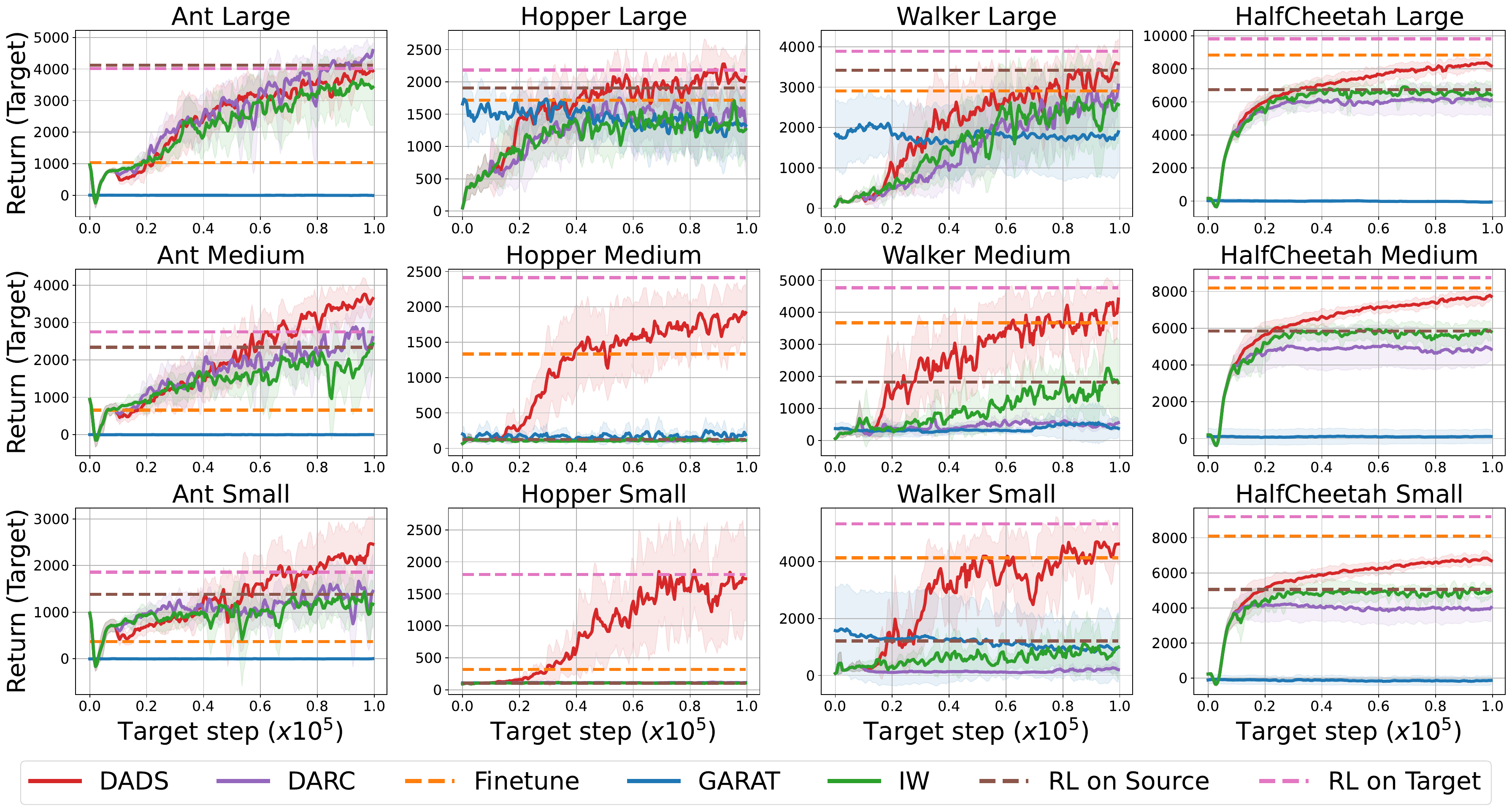}
    \caption{The target return of different methods in four Mujoco benchmarks with different deficient support levels: large overlapping support (Top row), medium overlapping support (Middle row), and small overlapping support (Bottom row). The solid curves are the average target returns over 5 runs with different random seeds, and the shaded areas represent standard deviation.}
    \label{fig:main_eval}
\end{figure*}

\textbf{Environments:} We use four simulated robot benchmarks from Mujoco Gym \cite{todorov2012mujoco, brockman2016openai}: Ant, HalfCheetah, Hopper and Walker. For each benchmark, we establish three levels of deficient support between the source and target domains: small overlapping, medium overlapping, and large overlapping. 
More specifically, for each benchmark, we first sample the noise $\xi$ from a pre-defined distribution $p_{\xi}$ and then add it to $s'$ as follows: 
\begin{equation}
    s' \leftarrow s' + \xi, \text{ where } \xi \sim p_{\xi}.
\end{equation}
We then update the value of the next state $s'$ in the environment. By adding noises from distinct distributions with different support sets, we simulate different levels of support deficiency between the source and the target domain. Specifically, the noise applied to the source domain only partially overlaps with the noise introduced to the target domain. Thus, after adding different noises to source and target, the source dynamics has the support deficiency w.r.t the target dynamics. Moreover, we adjust this overlapping region to create three different levels of support deficiency. 
Figure \ref{fig:ds_sample} illustrates the noises added to the source and the target domain at three deficient support levels in the Walker environment. Comprehensive details regarding the environments are provided in the \emph{Appendix}.

\textbf{Baselines:} We compare our algorithm with six baselines. \textit{DARC} \cite{eysenbach2020off}, which uses the additional reward term to encourage the policy to not use source transitions with low likelihood. \textit{GARAT} \cite{desai2020imitation} that learns the grounded source environment obtained via action transformation and then trains the policy on the learned environment. The \textit{Finetune} baseline first trains a policy on the source domain and then finetunes it with the limited transitions from the target domain. The \textit{IW} (Importance Weighting) baseline trains the policy on importance-weighted samples from the source domain. Finally, the \textit{RL on Target} trains the policy only using the target samples and can serve as Oracle. \textit{RL on Source} trains the policy only using source samples. We run all algorithms with the same five random seeds. More details about the baseline settings are in the \emph{Appendix}.

\subsection{Off-dynamics Policy Evaluation}
In Figure \ref{fig:main_eval}, we illustrate the off-dynamics policy performance of all methods across four Mujoco environments for the three support deficiency levels. Across all tasks, the performances of \textit{RL on Source} are considerably lower than \textit{RL on Target} performances, suggesting that directly transferring trained policies from the source to the target domain yields unsatisfactory performance when support deficiency is present. 

In cases with substantial support overlap (as seen in large overlap setting), the performance differences between the methods are not too high. However, as support overlap decreases (in medium and small overlap settings), our approach, DADS, consistently excels in most tasks. Specifically, \textit{GARAT} performances are significantly low for all tasks, supporting our intuition that its grounded action environment can be inaccurate and infeasible for policy learning under the dynamics mismatch and deficient support problems. The performances of \textit{DARC} and \textit{IW} drop significantly when the support overlap decreases (from a large level to a small level). Our method outperforms \textit{DARC} and \textit{IW} baselines for most of the tasks. We surpass the \textit{Finetune} baseline in nine out of twelve cases, excluding HalfCheetah. While \textit{Finetune} performs on par with our method on large and medium overlapping levels in HalfCheetah, it outperforms our method in the small overlap cases. We believe that this is because the agent never dies in the HalfCheetah environment, which helps it to transfer any learnings from the source domain to the target domain without any problem. Nonetheless, our method stands out as the only approach that asymptotically matches the performance of RL on Target (i.e. Oracle) and even surpasses \textit{RL on Target} in six out of twelve tasks.

\subsection{Ablation studies}
In this section, we analyze the impact of each component and hyperparameter in our method. We  provide the results for the Walker environment. The results for the other environments in all settings are provided in the \emph{Appendix}. 

\begin{figure}[ht]
    \centering
     \includegraphics[width=0.9\columnwidth]{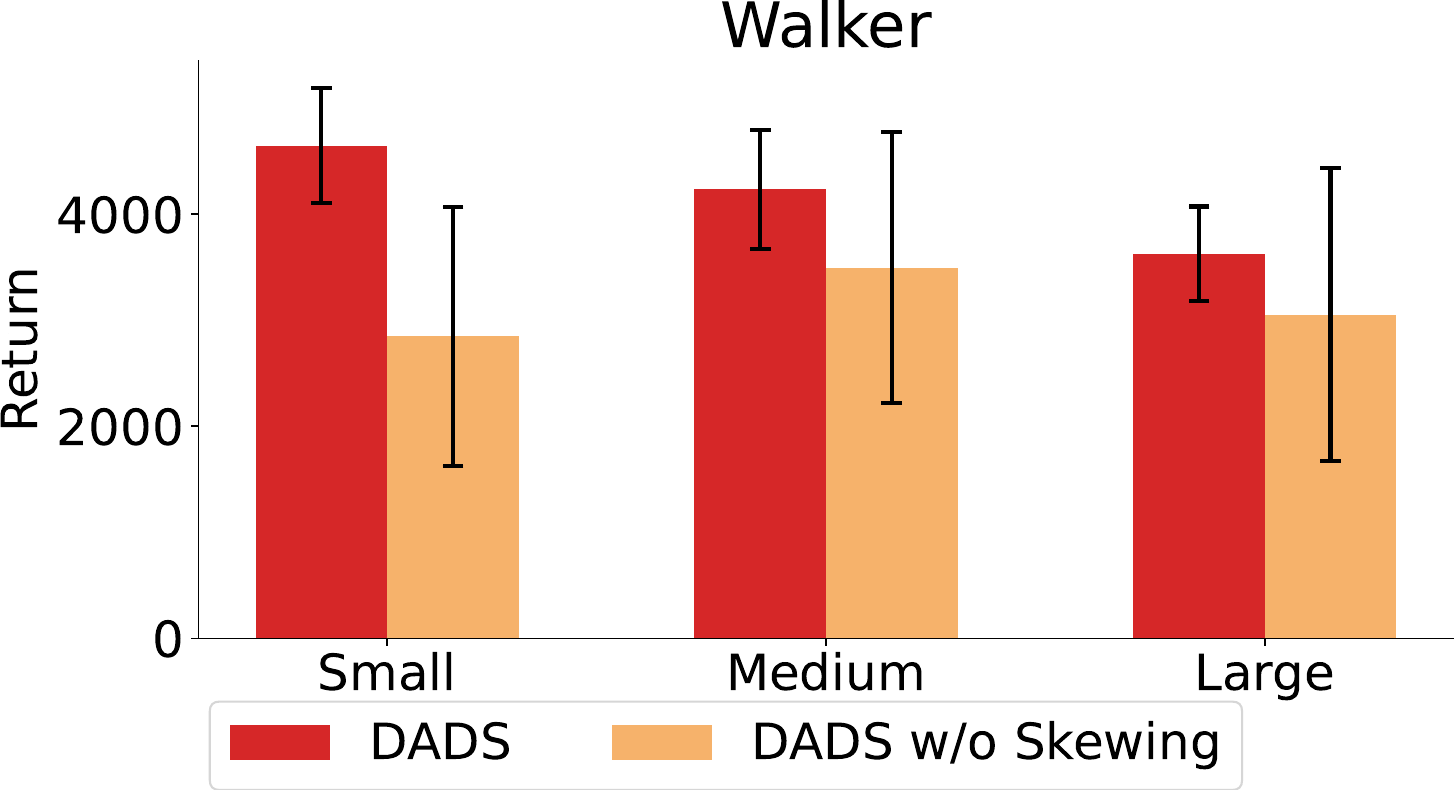}
    \caption{Comparison between our DADS method, and its variant with out Skewing opertation.}
    \label{fig:ablation_prb}
\end{figure}
\subsubsection{The impact of Skewing operation:}

To validate the effect of the skewing operation, we compared our method to a variant that does not use this component.
As shown in Figure \ref{fig:ablation_prb}, not including the skewing operation leads to a notable drop in performance and makes target return unstable, indicating the critical role of this component in our method.
\subsubsection{The impact of MixUp operation:}

\begin{figure}[ht]
    \centering
     \includegraphics[width=0.9\columnwidth]{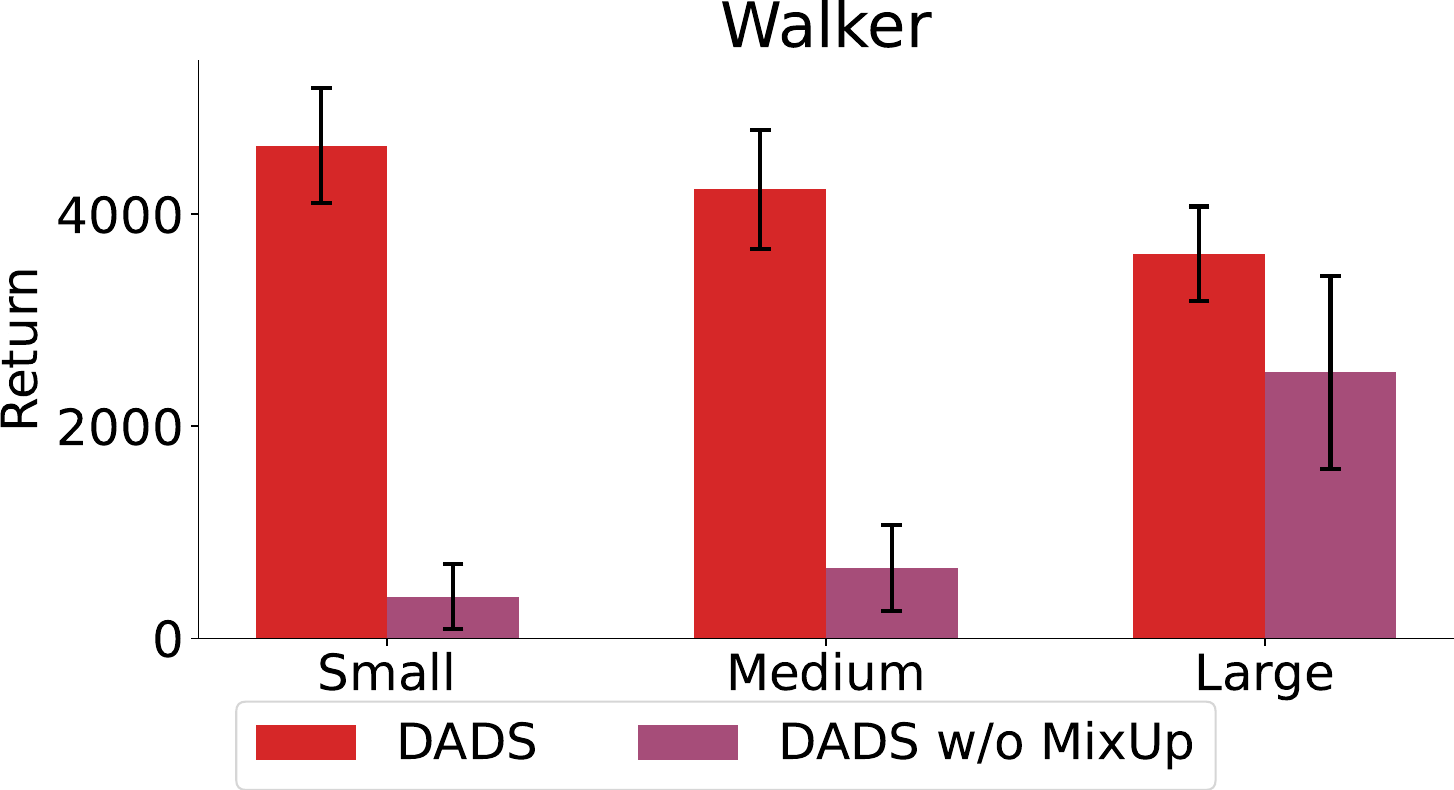}
    \caption{Comparison between our DADS method, and its variant without MixUp.}
    \label{fig:ablation_mixup}
\end{figure}

We evaluate the impact of the MixUp operation by removing it from our approach and retraining the policy. As shown in Figure \ref{fig:ablation_mixup}, excluding MixUp results in a significant performance drop in target return, especially in the settings where support deficiencies are large (medium and small overlap settings). This verifies the effectiveness of MixUp component in our method.

Notably, omitting either the Skewing or MixUp operations results in a significant reduction in target returns. This observation confirms the effectiveness of both operations.

\subsubsection{The impact of $\mu$:} 
\begin{figure}
    \centering
     \includegraphics[width=1.0\columnwidth]{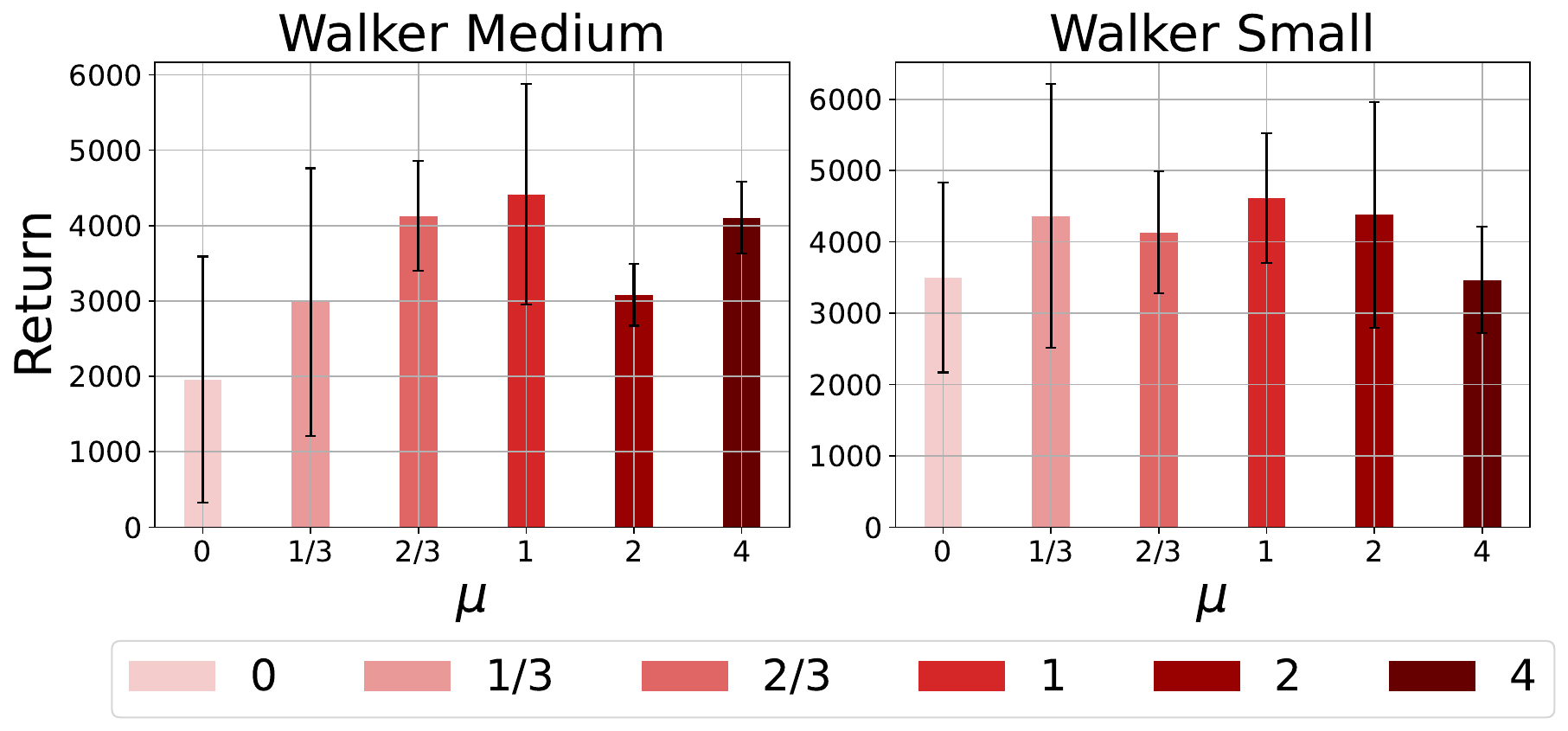}
    \caption{The adaptation performance of DADS with different values of $\mu$.}
    \label{fig:param_ablation}
\end{figure}
Our method only has one hyperparameter $\mu$ that controls the strength of the source dynamics regularization in Equation \ref{eq:dynamic_optimization_first}. We conduct experiments with different source dynamics regularization $\mu$ values  ($0, 1/3, 2/3, 1, 2, 4$), where $\mu = 0$ mean that we ignore the source dynamics regularization and increasing values of $\mu$ indicate higher weight for the source dynamics regularization. The results are shown in Figure \ref{fig:param_ablation}. We can see that the best range for $\mu$ is from $2/3$ to $1$. If we decrease the value of $\mu$ to $0$, the policy performance in the target domain drops significantly. The reason is that reducing or ignoring the source dynamics regularization results in sampling transitions that are too close to the target domain, which reduces the diversity of samples for the subsequent  MixUp operation. On the other hand, a high value of $\mu$ might also result in reduced, unstable performance as there are not enough source transitions that are close to the target domain. This could also adversely affect MixUp operations. Thus source dynamics constraint is important for the effective performance of our algorithm. 
In our experiments, we used $\mu = 1$ due to the highest target return values.
\section{Conclusion}
In this paper, we have addressed the problem of off-dynamics RL under deficient support, which is widely encountered in many real-world applications. To the best of our knowledge, ours is the first work on this problem. We proposed DADS, a simple, yet effective method, that reduces the support deficiency of the source domain by modifying it through two operations: skewing and extension. The skewing is learned by solving an optimization problem and the extension is performed by using a Mixup operation between source and target transitions. Extensive experiments have demonstrated the effectiveness of our method compared to the existing state-of-the-art approaches for off-dynamics RL.


\bibliographystyle{ACM-Reference-Format} 
\bibliography{AAMAS_2024}

\newpage
\onecolumn
\appendix

\section{Proof of Performance Bounds}
In this section, we provide the proof of the performance bounds in our main paper. We also present the telescoping lemma used for our theoretical results.
\progapfull*
\textit{Proof.} From the Lemma \ref{lemma:tele}, we have:
\begin{equation}
    \begin{aligned}
         &\eta_{\mathcal{M}_{tar}}(\pi) - \eta_{\mathcal{M}_{src}}(\pi)
        &= \frac{\gamma}{1 - \gamma}\cdot \mathbb{E}_{\rho_{tar}^{\pi}}\left[\int_{s'} P_{tar}(s'|s, a) V_{src}^\pi(s')ds' - P_{src}(s'|s, a) V_{src}^\pi(s')ds' \right]
    \end{aligned}
\end{equation}
So, we have
\begin{equation}
    \begin{aligned}
        \left|\eta_{\mathcal{M}_{tar}}(\pi) - \eta_{\mathcal{M}_{src}}(\pi)\right| 
        &= \frac{\gamma}{1 - \gamma}\cdot \left|\mathbb{E}_{\rho_{tar}^{\pi}}\left[\int_{s'} P_{tar}(s'|s, a) V_{src}^\pi(s')ds' - P_{src}(s'|s, a) V_{src}^\pi(s')ds' \right]\right| \\
        & \stackrel{(\mathrm{i})}{\leq} \frac{\gamma}{1 - \gamma} \cdot\mathbb{E}_{\rho_{tar}^\pi}\left[\int_{s'} \left|\left(P_{tar}(s'|s, a) - P_{src}(s'|s, a)\right) V_{src}^\pi\right| ds' \right] \\
        &\stackrel{(\mathrm{ii})}{\leq}\frac{\gamma r_{max}}{(1-\gamma)^2}\cdot\mathbb{E}_{\rho_{tar}^\pi}\left[\int_{s'}\left| P_{tar}(s'|s, a) - P_{src}(s'|s, a)\right| ds'\right] \\
        &\stackrel{(\mathrm{iii})}{=}\frac{2\gamma r_{max}}{(1-\gamma)^2}\cdot\mathbb{E}_{\rho_{tar}^\pi}\left[D_{TV}(P_{tar}(.|s, a), P_{src}(.|s, a))\right] \\
        &\stackrel{(\mathrm{iv})}{\leq}\frac{\gamma r_{max}}{(1-\gamma)^2}\cdot\sqrt{2\mathbb{E}_{\rho_{tar}^\pi}\left[D_{KL}(P_{tar}(.|s, a), P_{src}(.|s, a))\right]},
    \end{aligned}
\end{equation}
where (i) follows the absolute value of expectation inequality $\left|\mathbb{E}\left[X\right]\right| \leq \mathbb{E}\left[\left|X\right|\right]$, (ii) uses the bound of value function $V^\pi_{src}(s) \leq r_{max}/(1-\gamma)$, (iii) is from the total variation distance of probability measures, that  $\int_{s'}\left|P_{tar}(s'|s, a) - P_{src}(s'|s, a)\right|ds' = 2\cdot D_{TV}\left(P_{tar}(.|s, a)|P_{src}(.|s, a)\right)$, and (iv) directly applies the Pinsker's inequality.

\progapdefficient*

\textit{Proof.} Given $s \in \mathcal{S}$, $a \in \mathcal{A}$, and transition dynamics $P_{src}(.|s, a)$, $P_{tar}(.|s, a)$, we split the support set of the target dynamics $P_{tar}(.|s, a)$ into two distinct sets. The first set, $S_{s, a}^0 = \{[s'_{0_i}, s'_{0_j}]\}$, includes intervals where $P_{src}(.|s, a) = 0$. The second set, $S_{s, a}^1 = \{[s'_{1_i}, s'_{1_j}]\}$, includes intervals where $P_{src}(.|s, a) > 0$.
From the Lemma \ref{lemma:tele}, we have:
\begin{equation}
    \begin{aligned}
        \left|\eta_{tar}(\pi) - \eta_{src}(\pi)\right|
        &= \frac{\gamma}{1 - \gamma}\cdot\left|\mathbb{E}_{\rho_{tar}^\pi(s, a)}\left[\int_{s'}P_{tar}(s'|s, a)V_{src}^\pi(s') ds' - \int_{s'}P_{src}(s'|s, a)V_{src}^{\pi}(s, a) ds'\right] \right|\\
        &= \frac{\gamma}{1 - \gamma}\cdot\bigg|\mathbb{E}_{\rho_{tar}^\pi(s, a)}\bigg[\sum_{S_{s, a}^1}\int_{s'_{1_i}}^{s'_{1_j}}V_{src}^\pi(s') \cdot\big(P_{tar}(s'|s, a) - P_{src}(s'|s, a)\big)ds' 
        +\sum_{S_{s, a}^0}\int_{s'_{0_i}}^{s'_{0_j}}P_{tar}(s'|s, a)\cdot V_{src}^\pi(s') ds'\bigg]\bigg| \\
        &\stackrel{(\mathrm{i})}{\leq} \frac{\gamma}{1 - \gamma}\cdot\mathbb{E}_{\rho_{tar}^\pi(s, a)}\bigg[\bigg|\sum_{S_{s, a}^1}\int_{s'_{1_i}}^{s'_{1_j}} V_{src}^\pi(s')\cdot \big(P_{tar}(s'|s, a) - P_{src}(s'|s,a)\big)ds' 
        +\sum_{S_{s, a}^0}\int_{s'_{0_i}}^{s'_{0_j}}P_{tar}(s'|s, a)\cdot V_{src}^\pi(s')ds'\bigg|\bigg] \\
        &\stackrel{(\mathrm{ii})}{\leq} \frac{\gamma}{1 - \gamma}\cdot \mathbb{E}_{\rho_{tar}^\pi(s, a)}\bigg[\left|\sum_{S_{s, a}^1}\int_{s'_{1_i}}^{s'_{1_j}} V_{src}^\pi(s')\cdot \left(P_{tar}(s'|s, a) - P_{src}(s'|s,a) \right)ds'\right| \\
        &+\left|\sum_{S_{s, a}^0}\int_{s'_{0_i}}^{s'_{0_j}} P_{tar}(s'|s, a)\cdot V_{src}^\pi(s') ds'\right|\bigg] \\
        &\stackrel{(\mathrm{iii})}{\leq} \frac{\gamma}{1 - \gamma}\cdot \mathbb{E}_{\rho_{tar}^\pi(s, a)}\left[\sum_{S_{s, a}^1}\left|\int_{s'_{1_i}}^{s'_{1_j}} V_{src}^\pi(s')\cdot \left(P_{tar}(s'|s, a) - P_{src}(s'|s,a)\right)ds'\right|\right] \\
        &+\frac{\gamma}{1 - \gamma}\cdot \mathbb{E}_{\rho_{tar}^\pi(s, a)}\left[\sum_{S_{s, a}^0}\left|\int_{s'_{0_i}}^{s'_{0_j}}P_{tar}(s'|s, a)\cdot V_{src}^\pi(s')ds'\right|\right] \\
        &\stackrel{(\mathrm{iv})}{\leq} \frac{\gamma r_{max}}{(1-\gamma)^2} \cdot\mathbb{E}_{\rho_{tar}^\pi(s, a)}\left[\sum_{S_{s, a}^1}\left|\int_{s'_{1_i}}^{s'_{1_j}} P_{tar}(s'|s, a) - P_{src}(s'|s,a)ds'\right|\right] \\
        &+\frac{\gamma}{1 - \gamma}\cdot \mathbb{E}_{\rho_{tar}^\pi(s, a)}\left[\sum_{S_{s, a}^0}\int_{s'_{0_i}}^{s'_{0_j}} P_{tar}(s'|s, a)\cdot \left|V_{src}^\pi(s')\right|ds'\right],
    \end{aligned}
\end{equation}
where (i) uses the fact that $\left|\mathbb{E}\left[X\right]\right| \leq \mathbb{E}\left[\left|X\right|\right]$, (i) and (iii) are from the triangle inequalities $\left|x + y\right| \leq \left|x\right| +\left|y\right|$ and $\left|\sum_{i=1}^N(x_i)\right| \leq \sum_{i=1}^N\left(\left|x_i\right|\right)$, and (iv) is from the bound of value function $V^\pi_{src}(s) \leq r_{max}/(1-\gamma)$.

\begin{lemma}[Telescoping Lemma, Lemma 4.3 in \cite{luo2018algorithmic}]\label{lemma:tele}
Let $\mathcal{M}_1 := (\mathcal{S}, \mathcal{A}, P_1, r, \gamma, \rho_0) $ and $\mathcal{M}_2 := (\mathcal{S}, \mathcal{A}, P_2, r, \gamma, \rho_0)$ be two MDPs with different dynamcis $P_1$ and $P_2$. Given a policy $\pi$, we have
\begin{equation}
    \begin{aligned}
        &\eta_{\mathcal{M}_1}(\pi) - \eta_{\mathcal{M}_2}(\pi) \\
        &=\frac{\gamma}{1 - \gamma}\mathbb{E}_{s, a \sim \rho_{\mathcal{M}_1}^\pi}\left[\mathbb{E}_{s^{\prime} \sim P_1}\left[V_{\mathcal{M}_2}^\pi\left(s^{\prime}\right)\right]-\mathbb{E}_{s^{\prime} \sim P_2}\left[V_{\mathcal{M}_2}^\pi\left(s^{\prime}\right)\right]\right]
    \end{aligned}
\end{equation}
\end{lemma}


\section{Derivation of Skewing Source Dynamics}
In the Skewing source dynamics, we find the dynamics function $P(s'|s, a)$ that is closely aligned with the target dynamics $P_{tar}(s'|s, a)$ but not significantly far from the source dynamics $P_{src}(s'|s, a)$, measured in terms of the KL divergence. This is formulated as the subsequent constrained function optimization problem:
\begin{equation}
\begin{aligned} \label{eq:dynamics_optimization_first}
    &\min_{P \in \mathcal{P}} D_{\mathrm{KL}}(P(.|s,a)||P_{tar}(.|s,a)) \text{ } \forall s \in \mathcal{S}, a \in \mathcal{A}. \\
    & \text{s.t } D_{\mathrm{KL}}(P(.|s,a)||P_{src}(.|s,a)) \leq \epsilon \text{ } \forall s \in \mathcal{S}, a \in \mathcal{A}. \\
    & \int_{s'} P(s'|s, a)ds' = 1 \text{ } \forall s \in \mathcal{S}, a \in \mathcal{A}.
\end{aligned}    
\end{equation}
where $\mathcal{P}$ is a family of all transition dynamics distributions, and we have $P_{tar}(.|s,a)\in \mathcal{P}$ and $P_{src}(.|s,a)\in \mathcal{P}$. The first constraint with KL divergence and the parameter $\epsilon$ regularizes the dynamics function $P(s'|s,a)$ to stay close to the source dynamics $P_{src}(s'|s, a)$, while the second constraint is to ensure $P(s'|s, a)$ is a valid probability density function. Since imposing the pointwise KL in the above objective across all states and actions is computationally challenging, we relax by enforcing them only in expectation with $\rho^{\pi}_{src}(s,a)$ is the state-action distribution in the source domain \( \mathcal{M}_{src} \). Thus, we obtain:
\begin{equation}
\begin{aligned} \label{eq:dynamics_optimization}
    &\min_{P \in \mathcal{P}} \int_{s,a} \rho^{\pi}_{src}(s,a)D_{\mathrm{KL}}(P(.|s,a)||P_{tar}(.|s,a))dsda. \\
    & \text{s.t } \int_{s,a} \rho^{\pi}_{src}(s,a) D_{\mathrm{KL}}(P(.|s,a)||P_{src}(.|s,a))dsda \leq \epsilon. \\
    & \int_{s'} P(s'|s, a)ds' = 1 \text{ } \forall s \in \mathcal{S}, a \in \mathcal{A}.
\end{aligned}  
\end{equation}
Similar with \cite{peng2019advantage}, from (\ref{eq:dynamics_optimization}), we form the Lagrangian as follows:
\begin{equation}
\begin{aligned}
    \mathcal{L}(P, \mu, \upsilon) &= \int_{s,a} \rho^{\pi}_{src}(s,a)D_{\mathrm{KL}}(P(.|s,a)||P_{tar}(.|s,a))dsda + \mu[\int_{s,a}\rho^{\pi}_{src}(s,a)D_{\mathrm{KL}}(P(.|s,a)||P_{src}(.|s,a))dsda - \epsilon] \\ &+ \int_{s,a}\upsilon_{sa}[1 - \int_{s'}P(s'|s,a)ds']dsda .
\end{aligned}
\end{equation}
where $\mu$ and $\upsilon = \{\upsilon_{sa} | \forall s \in \mathcal{S}, a \in \mathcal{A}\}$ corresponds to the Lagrange multipliers. Differentiating $\mathcal{L}(P, \mu, \upsilon)$ with respect to $P(s'|s,a)$ we have:
\begin{equation}
\begin{aligned}
    \frac{\partial\mathcal{L}}{\partial P}  
    &= \rho^{\pi}_{src}(s,a)[\log P(s'|s,a) - \log P_{tar}(s'|s,a) + 1] 
    + \mu\rho^{\pi}_{src}(s,a)[\log P(s'|s,a) - \log P_{src}(s'|s,a) + 1] 
    - \upsilon_{sa}. 
    \\
    &= \log P(s'|s,a) \rho^{\pi}_{src}(s,a)(1+\mu) + \rho^{\pi}_{src}(s,a)(1+\mu) -\rho^{\pi}_{src}(s,a)[\log P_{tar}(s'|s,a) + \mu \log P_{src}(s'|s,a) ] -\upsilon_{sa}.
\end{aligned}
\end{equation}
Setting to zero and solving for $P(s'|s, a)$ gives:
\begin{equation}
\begin{split}
    \log P(s'|s,a) = \frac{1}{1 + \mu}[\log P_{tar}(s'|s,a) +\mu\log P_{src}(s'|s,a)] + \frac{\upsilon_{sa}}{\rho^{\pi}_{src}(s,a)(1 + \mu)} - 1.
\end{split}
\end{equation}
or 
\begin{equation}\label{eq:p_optimal_1}
\begin{split}
    P(s'|s,a) = P_{src}(s'|s,a)\exp[\frac{1}{1 + \mu}(\log P_{tar}(s'|s,a) -\log P_{src}(s'|s,a))]\exp[\frac{\upsilon_{s,a}}{\rho^{\pi}_{src}(s,a)(1 + \mu)} -1].
\end{split}
\end{equation}
Since $\int_{s'}P(s'|s, a)ds' = 1$, the second exponential term in the Eq (\ref{eq:p_optimal_1}) is the partition function $Z(s, a)$ that normalizes the condition state action distribution as follows:
\begin{equation}
\begin{split}
    Z(s,a) &= \exp[1 - \frac{\upsilon_{s,a}}{ \rho^{\pi}_{src}(s,a)(1 + \mu)}] \\ &= \text{ }\int_{s'} P_{src}(s'|s,a)\exp[\frac{1}{1 + \mu}\log \frac{P_{tar}(s'|s,a)}{P_{src}(s'|s,a)}]ds'.
\end{split}
\end{equation}
Finally, the optimal $P^*(s'|s, a)$ is given by:
\begin{equation}
\begin{aligned}
    P^*(s'|s,a) &\propto P_{src}(s'|s, a)\exp[\frac{1}{1 + \mu}\log \frac{P_{tar}(s'|s,a)}
    {P_{src}(s'|s,a)}].
\end{aligned}
\end{equation}
\section{Experimental Details and Hyperparameters}
We use Python 3.9, Pytorch 1.13.1, and Gym 0.23.1. All experiments are conducted on an Ubuntu 20.4 server with 36 cores CPU, 767GB RAM, and a V100 32GB GPU with CUDA version 12.0. Our source code is provided in the supplementary material and will be made publicly accessible upon publication.
\subsection{Environments}
\begin{table*}
\centering
\begin{tabular}{@{}cclll@{}}
\toprule
            &        & \multicolumn{1}{c}{Large overlapping level}                                                                 & \multicolumn{1}{c}{Medium overlapping level}                                                                & \multicolumn{1}{c}{Small overlapping level}                                                                 \\ \midrule
Ant         & Source & \begin{tabular}[c]{@{}l@{}}$p_{\xi_{pos}}$: $\mathcal{U}(-0.01, 0.01)$\\ $p_{\xi_{vel}}$: $\mathcal{N}(0.0, 0.005)$\end{tabular} & \begin{tabular}[c]{@{}l@{}}$p_{\xi_{pos}}$: $\mathcal{U}(-0.01, 0.01)$\\ $p_{\xi_{vel}}$: $\mathcal{N}(0.0, 0.005)$\end{tabular} &  \begin{tabular}[c]{@{}l@{}}$p_{\xi_{pos}}$: $\mathcal{U}(-0.01, 0.01)$\\ $p_{\xi_{vel}}$: $\mathcal{N}(0.0, 0.005)$\end{tabular}                                                                       \\
            & Target & \begin{tabular}[c]{@{}l@{}}$p_{\xi_{pos}}$: $\mathcal{U}(-0.005, 0.015)$\\ $p_{\xi_{vel}}$: $\mathcal{N}(-0.0125, 0.005)$\end{tabular} & \begin{tabular}[c]{@{}l@{}}$p_{\xi_{pos}}$: $\mathcal{U}(0.0, 0.02)$\\ $p_{\xi_{vel}}$: $\mathcal{N}(-0.015, 0.005)$\end{tabular}  & \begin{tabular}[c]{@{}l@{}}$p_{\xi_{pos}}$: $\mathcal{U}(0.005, 0.025)$\\ $p_{\xi_{vel}}$: $\mathcal{N}(-0.0175, 0.005)$\end{tabular} 
                     \\
Hopper      & Source & \begin{tabular}[c]{@{}l@{}}$p_{\xi_{pos}}$: $\mathcal{U}(-0.01, 0.01)$\\ $p_{\xi_{vel}}$: $\mathcal{N}(0.0, 0.005)$\end{tabular} & \begin{tabular}[c]{@{}l@{}}$p_{\xi_{pos}}$: $\mathcal{U}(-0.01, 0.01)$\\ $p_{\xi_{vel}}$: $\mathcal{N}(0.0, 0.005)$\end{tabular} &  \begin{tabular}[c]{@{}l@{}}$p_{\xi_{pos}}$: $\mathcal{U}(-0.01, 0.01)$\\ $p_{\xi_{vel}}$: $\mathcal{N}(0.0, 0.005)$\end{tabular}                                                                       \\
            & Target & \begin{tabular}[c]{@{}l@{}}$p_{\xi_{pos}}$: $\mathcal{U}(-0.005, 0.015)$\\ $p_{\xi_{vel}}$: $\mathcal{N}(-0.0125, 0.005)$\end{tabular} & \begin{tabular}[c]{@{}l@{}}$p_{\xi_{pos}}$: $\mathcal{U}(0.0, 0.02)$\\ $p_{\xi_{vel}}$: $\mathcal{N}(-0.015, 0.005)$\end{tabular}  & \begin{tabular}[c]{@{}l@{}}$p_{\xi_{pos}}$: $\mathcal{U}(0.005, 0.025)$\\ $p_{\xi_{vel}}$: $\mathcal{N}(-0.0175, 0.005)$\end{tabular}                                                                                                   \\
Walker      & Source & \begin{tabular}[c]{@{}l@{}}$p_{\xi_{pos}}$: $\mathcal{U}(-0.02, 0.02)$\\ $p_{\xi_{vel}}$: $\mathcal{U}(-0.02, 0.02)$\end{tabular} & \begin{tabular}[c]{@{}l@{}}$p_{\xi_{pos}}$: $\mathcal{U}(-0.02, 0.02)$\\ $p_{\xi_{vel}}$: $\mathcal{U}(-0.02, 0.02)$\end{tabular} & \begin{tabular}[c]{@{}l@{}}$p_{\xi_{pos}}$: $\mathcal{U}(-0.02, 0.02)$\\ $p_{\xi_{vel}}$: $\mathcal{U}(-0.02, 0.02)$\end{tabular}
                     \\
            & Targe  & \begin{tabular}[c]{@{}l@{}}$p_{\xi_{pos}}$: $\mathcal{N}(0.015, 0.004)$\\ $p_{\xi_{vel}}$: $\mathcal{N}(-0.015, 0.004)$\end{tabular}   & \begin{tabular}[c]{@{}l@{}}$p_{\xi_{pos}}$: $\mathcal{N}(0.02, 0.004)$\\ $p_{\xi_{vel}}$: $\mathcal{N}(-0.02, 0.004)$\end{tabular} & \begin{tabular}[c]{@{}l@{}}$p_{\xi_{pos}}$: $\mathcal{N}(0.025, 0.004)$\\ $p_{\xi_{pos}}$: $\mathcal{N}(-0.025, 0.004)$\end{tabular} 
                     \\
HalfCheetah & Source & \begin{tabular}[c]{@{}l@{}}$p_{\xi_{pos}}$: $\mathcal{U}(-0.02, 0.02)$\\ $p_{\xi_{vel}}$: $\mathcal{U}(-0.02, 0.02)$\end{tabular} & \begin{tabular}[c]{@{}l@{}}$p_{\xi_{pos}}$: $\mathcal{U}(-0.02, 0.02)$\\ $p_{\xi_{vel}}$: $\mathcal{U}(-0.02, 0.02)$\end{tabular} & \begin{tabular}[c]{@{}l@{}}$p_{\xi_{pos}}$: $\mathcal{U}(-0.02, 0.02)$\\ $p_{\xi_{vel}}$: $\mathcal{U}(-0.02, 0.02)$\end{tabular} 
                     \\
            & Target & \begin{tabular}[c]{@{}l@{}}$p_{\xi_{pos}}$: $\mathcal{N}(-0.015, 0.004)$\\ $p_{\xi_{vel}}$: $\mathcal{N}(-0.015, 0.004)$\end{tabular}   & \begin{tabular}[c]{@{}l@{}}$p_{\xi_{pos}}$: $\mathcal{N}(-0.02, 0.004)$\\ $p_{\xi_{vel}}$: $\mathcal{N}(-0.02, 0.004)$\end{tabular} & \begin{tabular}[c]{@{}l@{}}$p_{\xi_{pos}}$: $\mathcal{N}(-0.025, 0.004)$\\ $p_{\xi_{vel}}$: $\mathcal{N}(-0.025, 0.004)$\end{tabular}  \\ \bottomrule
\end{tabular}
\caption{Details of noise distributions for all environments. We note $\mathcal{U}(a, b)$ is the Uniform distribution with two boundaries parameters $a \text{ and } b$, and $\mathcal{N}(m,s)$ is the Gaussian distribution with mean is $m$ and standard deviation $s$.}
\label{tab: env_config}
\end{table*}
\begin{figure}[H]
    \centering
    \includegraphics[width=0.55\linewidth]{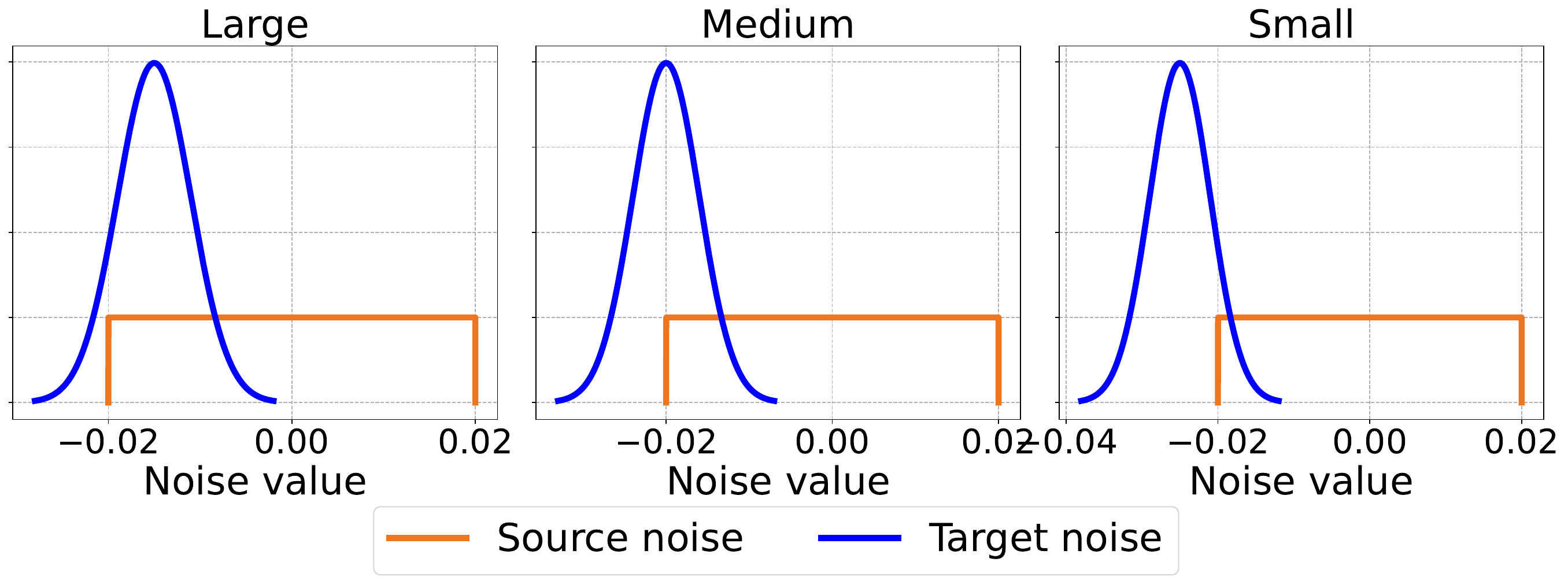}
    \caption{Visualization of the source and target noise distributions in the Walker benchmark in three distinct deficient support levels.}
    \label{fig:ds_sample}
\end{figure}
We use four simulated robot benchmarks from Mujoco Gym \cite{todorov2012mujoco, brockman2016openai}: Ant-v2, Hopper-v2, Walker-v2, and HalfCheetah-v2. For each benchmark, we establish three levels of deficient support between the source and target domains: small, medium, and large. 
More specifically, for each benchmark, we first sample the noise $\xi$ from a pre-defined distribution $p_{\xi}$ and then add it to $s'$ as follows: 
\begin{equation*}
    s' \leftarrow s' + \xi, \text{ where } \xi \sim p_{\xi}
\end{equation*}
We then update the value of the next state $s'$ in the environment. By adding noises from distinct distributions with different support sets, we simulate different levels of support deficiency between the source and the target domain. Specifically, the noise applied to the source domain only partially overlaps with the noise introduced to the target domain. Thus, after adding different noises to source and target, the source dynamics has the support deficiency w.r.t the target dynamics. Moreover, we adjust this overlapping region to create three different levels of support deficiency. Figure \ref{fig:ds_sample} visualizes the noises added to the source and the target domain at three deficient support levels in the Walker environment. Note that the next state $s'$ is a vector of position feature $s'_{pos}$ and velocity feature $s'_{vel}$. We use the noise distribution $p_{\xi_{pos}}$ and $p_{\xi_{vel}}$ for each postion feature $s'_{pos}$ and velocity feature $s'_{vel}$ respectively.
A sample implementation of the environment is shown in Example \ref{tab: env_config}.
\renewcommand{\lstlistingname}{Example}
\begin{lstlisting}[ 
language=Python, 
label=code:NoiseEnv,
captionpos=b,
aboveskip=1em,
belowskip=1em,
numberstyle=\tiny,
caption=A sample Python pseudo implementation of the Walker environment.]
import numpy as np
from gym import utils
from gym.envs.mujoco import mujoco_env

class Walker2dNoiseEnv(mujoco_env.MujocoEnv, utils.EzPickle):
    def __init__(self, pos_noise_dis, vel_noise_dis):
        self.pos_noise_dis = pos_noise_dis
        self.vel_noise_dis = vel_noise_dis
        mujoco_env.MujocoEnv.__init__(self, "agent.xml", 4) 
        utils.EzPickle.__init__(self)

    def step(self, a):
        posbefore = self.sim.data.qpos[0]
        self.do_simulation(a, self.frame_skip)
        # adding noise to the qpos and qvel
        qpos_noise = self.pos_noise_dis(size=self.model.nq)
        qvel_noise = self.vel_noise_dis(size=self.model.nv)
        qpos = self.sim.data.qpos + qpos_noise
        qvel = self.sim.data.qvel + qvel_noise
        # set the new qpos and qvel after adding noise
        self.set_state(qpos, qvel)
        # calculate the reward based on the state, next state and action
        posafter, height, ang = self.sim.data.qpos[0:3]
        alive_bonus = 1.0
        reward = (posafter - posbefore) / self.dt
        reward += alive_bonus
        reward -= 1e-3 * np.square(a).sum()
        done = not (height > 0.8 and height < 2.0 and ang > -1.0 and ang < 1.0)
        ob = self._get_obs()
        return ob, reward, done, {}

        

        



        


        
        
\end{lstlisting}
We provide the details of noise distributions for all environments in Table \ref{tab: env_config}.

\subsection{Algorithm's Implementations}
This section provides the implementation details and hyperparameters of all methods.  
\paragraph{DADS} 
We use Soft Actor-Critic (SAC) \cite{haarnoja2018soft} as the RL algorithm for policy learning. Our setup is based on SAC's implementation by \cite{huang2022cleanrl}. We use three-layer MLPs with 256 hidden units for policy and value functions. We set the learning rate at $3e^{-4}$ and batch size at $128$. The discount factor $\gamma$ is set to $0.99$ in all environments. The target smoothing coefficient $\tau$ is 0.005. The initial temperature coefficient is $1.0$ and is learnable. We set the gradient step for the policy as 1, and the target update interval is $1$. The buffer size is $10^6$. For each domain classifier, we employ two-layer MLPs with 256 hidden units. We use the Maximum cross-entropy loss to train all of the domain classifiers. As suggested in \cite{eysenbach2020off}, we add the standard Gaussian noise to the input of the classifiers during the classifier's training. We warm-started \emph{DADS} by applying RL on the source task without modifying the reward for the first $10^5$ iterations. The implementation of the sampling method in the Skewing operation is based on \cite{schaul2015prioritized} implemented in \cite{labml}. We train our algorithm with $10^6$ steps and collect data from the target domain after $10$ steps, i.e., the ratio $r$ of experience from source vs. target is 10.
\paragraph{DARC} We follow the implementation of \emph{DARC} algorithm from the official code\footnote[1]{\url{https://github.com/google-research/google-research/tree/master/darc}}. We run \emph{DARC} with the default configuration reported in the paper \cite{eysenbach2020off} and the same SAC backbone as \emph{DADS}. We train \emph{DARC} with $10^6$ steps and collect target data after $10$ steps.
\paragraph{GARAT} We utilize the author's official \emph{GARAT} implementation from their GitHub repository\footnote[2]{\url{https://github.com/HareshKarnan/GARAT-NeurIPS20}}. We integrate our environments into the code and execute it using the default hyperparameters mentioned in the paper \cite{desai2020imitation}. For a fair comparison with other methods, we increase the target transitions in \emph{GARAT} to $10^5$ for learning the grounded source environment and then train the policy with $10^5$ interactions with the grounded source environments. 
\paragraph{RL on Target and RL on Source} Two baselines are implemented with the same SAC backbone with our method \emph{DADS}. The only difference is that \emph{RL on Target} is trained with the target domain, while \emph{RL on Source} uses the source domain. Both baselines are trained with $10^6$ steps. 
\paragraph{Importance weight} The \emph{Importance Weight (IW)} baseline calculates importance weights by approximating $P_{tar}(s'|s, a) / P_{src}(s'|s, a)$ as $\exp(\Delta r)$. These weights are then used to weight transitions in the SAC actor and critic losses.
\paragraph{Finetune} We first train the SAC on the source domain with $10^6$ environment interactions. Then, we transfer SAC to the target domain and continue the training with $10^5$ target interactions.

Following \cite{fujimoto2021minimalist}, we evaluate all methods every 5000 time steps during training, with each evaluation comprising 10 episodes.

\section{Additional Experiments}
In this section, we provide the additional results for ablation studies in all tasks.

\subsection{The impact of Skewing operation}
We assess the effect of the Skewing operation across all tasks. As shown in Figure \ref{fig:skew_mixup_full}, the absence of skewing operation results in a notable performance drop in target return in almost environments, indicating the importance of this component in our method.

\subsection{The impact of MixUp operation}
We evaluate the impact of the MixUp operation by removing it from our approach and retraining the policy. As shown in Figure \ref{fig:skew_mixup_full}, excluding MixUp results in a significant performance drop in target return. This observation verifies the effectiveness of the MixUp component in our method.

\subsection{The impact of $\mu$} 
\begin{figure*}[ht]
    \centering
    \includegraphics[width=1.0\linewidth]{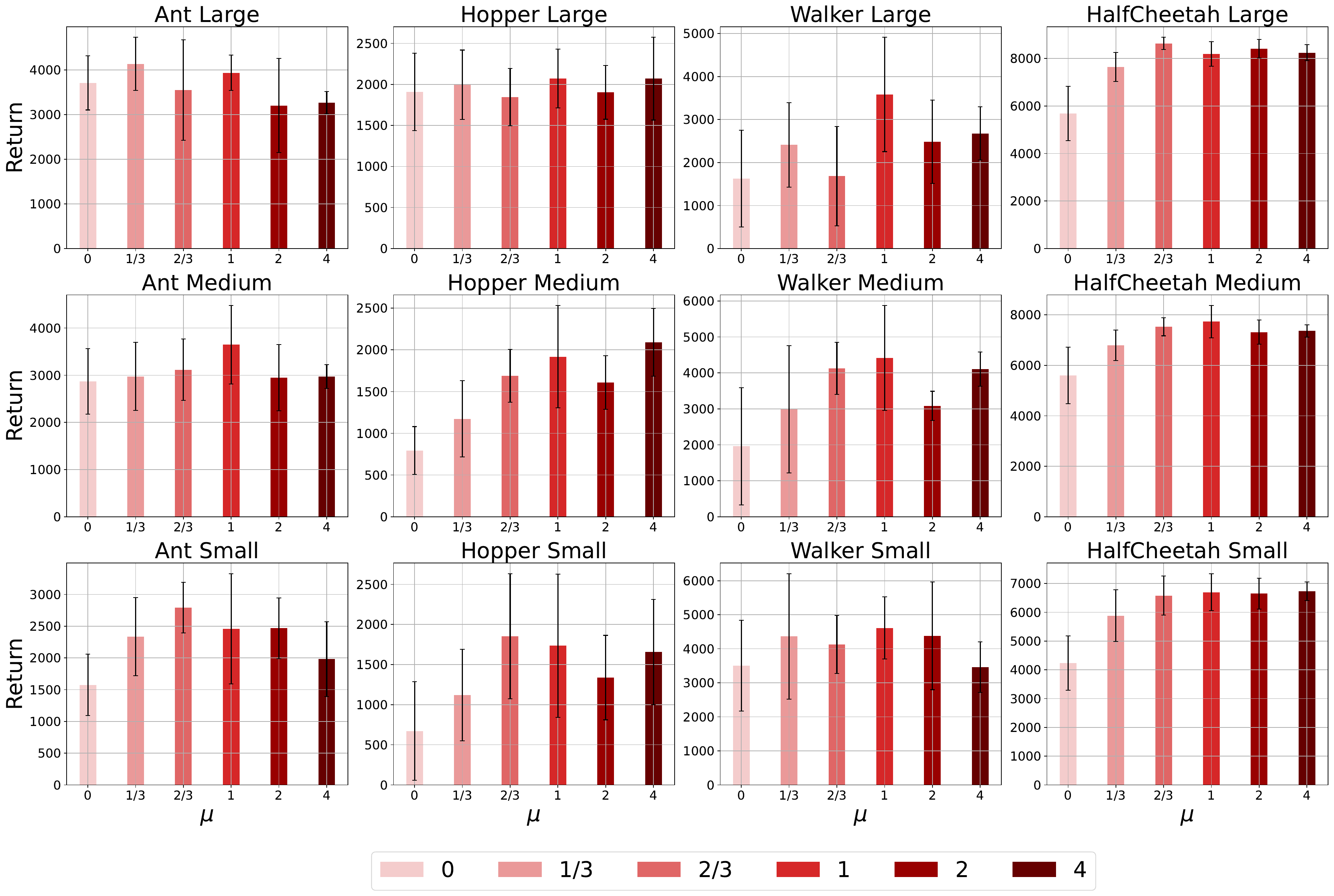}
    \caption{The adaptation performance of DADS with different values of $\mu$ in all tasks.}
    \label{fig:hyperparam}
\end{figure*}

Figure \ref{fig:hyperparam} illustrates the results of an ablation study about the impact of hyperparameter $\mu$, which controls the weight for the source dynamics regularization. We used $\mu$ = 1 in our experiments due to the highest target return values.

\begin{figure*}[ht]
    \centering
    \includegraphics[width=1.0\linewidth]{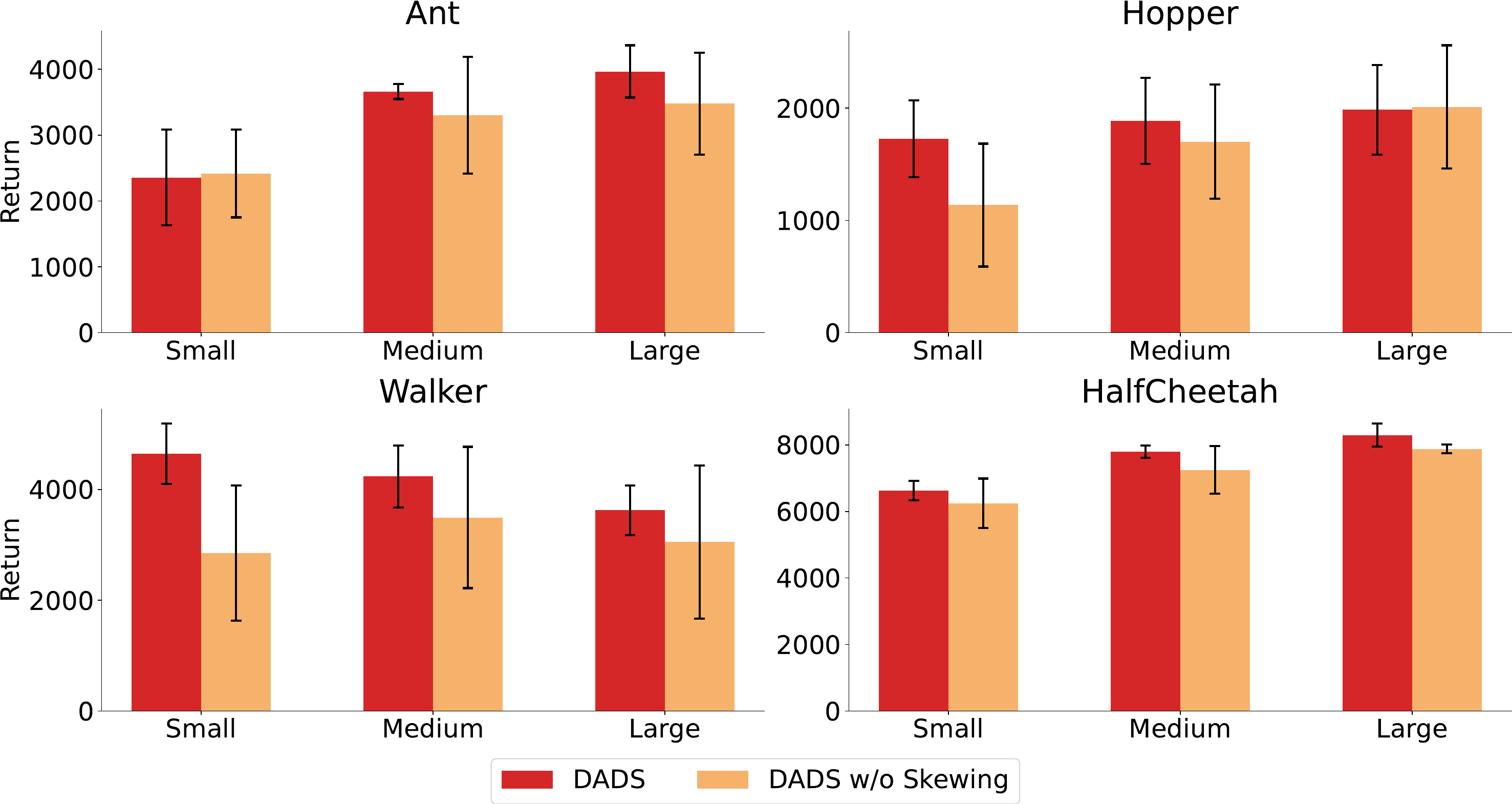}
    \caption{Comparison between our DADS method and its variant without Skewing operation in all environments.}
    \label{fig:skew_mixup_full}
\end{figure*}

\begin{figure*}[ht]
    \centering
    \includegraphics[width=1.0\linewidth]{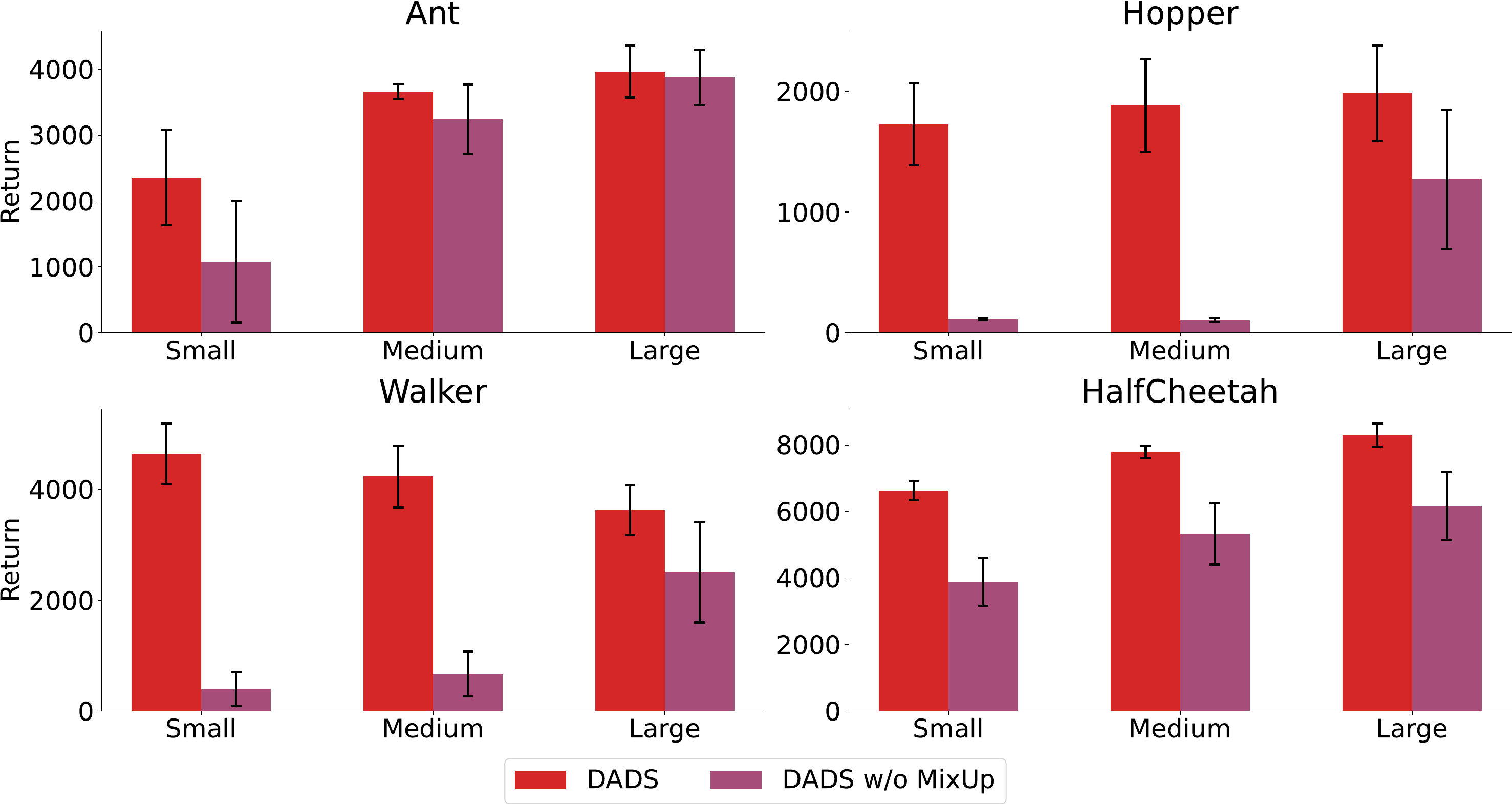}
    \caption{Comparison between our DADS method and its variant without using MixUp in all environments.}
    \label{fig:skew_mixup_full}
\end{figure*}

\end{document}